\documentclass[10pt,twocolumn,letterpaper]{article}

\usepackage[pagenumbers]{wacv} 

\usepackage{graphicx}
\usepackage{amsmath}
\usepackage{amssymb}
\usepackage{booktabs}
\usepackage{multirow}
\usepackage{array}
\usepackage{tabularx}
\usepackage{pifont}
\usepackage{xspace}
\usepackage{tikz}
\usetikzlibrary{positioning,arrows.meta,fit,calc}
\usepackage{microtype}
\usepackage{pgfplots}
\pgfplotsset{compat=1.17}

\newcommand{\cmark}{\ding{51}}
\newcommand{\pmark}{$\circ$}
\newcommand{\xmark}{--}
\newcommand{\usr}{USR\xspace}

\definecolor{wacvblue}{rgb}{0.21,0.49,0.74}
\usepackage[pagebackref,breaklinks,colorlinks,allcolors=wacvblue]{hyperref}

\def\wacvPaperID{*****} 
\def\confName{WACV}
\def\confYear{2027}

\title{Urban State Representation for Autonomous Mobility: A Survey of\\
Multimodal Perception, Architectures, Datasets, and Benchmarks}

\author{Binay Kumar Singh \qquad Niels Da Vitoria Lobo\\
University of Central Florida, Orlando, FL, USA\\
{\tt\small binay.singh@ucf.edu \qquad Niels.DaVitoriaLobo@ucf.edu}
}

\begin{document}
\maketitle
\begin{abstract}
Cities are sensed by many platforms at once: on-board cameras, LiDAR and radar on autonomous vehicles, roadside units at intersections, drones and satellites overhead, and the GNSS traces of millions of trips. Yet most learned representations are built for one platform, one city, and one task. We survey this fragmented landscape through a single lens, the \emph{Urban State Representation} (\usr): a georeferenced, layered, time-varying, uncertainty-aware state that any sensing platform can write to and any downstream task can read from. We (i)~formalize \usr and five desiderata for it, and relate it to bird's-eye-view (BEV) features, occupancy, vectorized maps, neural fields and 3D Gaussians, scene graphs, road-network graphs, language tokens, and world-model latents; (ii)~organize sensing platforms, cross-view (ground--aerial) matching, fusion operators, 3D geometric foundation models, end-to-end and vision-language-action models, and generative world models and data engines in one taxonomy that spans sensors to decisions; (iii)~compile 26 urban datasets and simulators across ego, cooperative (V2X), aerial, trajectory, and language settings, with a metric glossary, a cross-benchmark results snapshot, a trend analysis, and a platform-coverage gap analysis; and (iv)~review applications and propose, as an outlook, \emph{USR-5}, a five-test protocol with explicit references and one aggregate score for measuring whether a state is shared, together with a research agenda of six open problems covering cross-city and cross-sensor transfer, uncertainty, efficiency, sim-to-real evaluation, and governance. This survey is an entry point for computer vision, autonomous driving, transportation, and urban computing researchers; the paper list, datasets, and USR-5 code are at \url{https://github.com/binayksingh/Awesome-Urban-State-Representation}.
\end{abstract}

\section{Introduction}
\label{sec:intro}

Autonomous mobility in cities depends on a model of the world that is accurate, current, and actionable. The sensing needed to build such a model is now abundant. A single autonomous test vehicle carries multiple cameras, LiDARs and radars~\cite{caesar2020nuscenes,sun2020waymo}. Instrumented intersections stream roadside camera and LiDAR data~\cite{yu2022dairv2x,zimmer2024tumtraf}. Drones capture naturalistic trajectories from above~\cite{krajewski2018highd,zhu2022visdrone}. Loop detectors and phones record city-scale mobility~\cite{li2018dcrnn,zheng2010geolife}.

The \emph{representations} learned from this data, however, remain fragmented. Perception builds BEV or occupancy features for one ego vehicle~\cite{li2022bevformer,tian2023occ3d}; cooperative perception fuses features across a few agents~\cite{xu2022v2xvit,hu2022where2comm}; neural rendering reconstructs scenes for simulation~\cite{tancik2022blocknerf,yan2024streetgaussians}; and urban computing learns from traces and maps with little contact with sensor-level perception~\cite{zheng2014urbancomputing,zhang2024ufmsurvey}. Each community has its own datasets, metrics, and vocabulary, even though all describe the same streets at the same moments.

\noindent\textbf{Scope and positioning.} Strong surveys cover individual pieces: BEV perception~\cite{li2024bevsurvey}, occupancy~\cite{xu2025occsurvey}, collaborative and V2X perception~\cite{han2023collabsurvey,huang2025v2xcp}, end-to-end driving~\cite{chen2024e2esurvey}, multimodal LLMs for driving~\cite{cui2024mllmsurvey}, Gaussian splatting~\cite{chen2024gssurvey}, driving world models~\cite{feng2025wmsurvey,tu2026dwmsurvey}, and urban foundation models~\cite{zhang2024ufmsurvey}. Closest to our framing, Hu \etal~\cite{hu2026maead} review shared world models maintained across vehicles and infrastructure; we extend the shared-state view to aerial platforms and mobility traces, connect it to urban computing, and propose a protocol to measure sharing. As \cref{tab:surveys} shows, none spans every sensing platform \emph{and} the full chain from sensing to evaluation.

\noindent\textbf{Survey methodology.} This is a structured narrative survey rather than a systematic review. We focus on 2017--2026 work, plus earlier foundational datasets and methods, at vision, learning, robotics, transportation, and data-mining venues (CVPR, ICCV, ECCV, WACV, NeurIPS, ICLR, ICRA, CoRL, ITSC, KDD) plus influential preprints, seeded by the surveys in \cref{tab:surveys} and expanded by citation search and by searching these venues for each family and platform (\eg, BEV, occupancy, V2X, HD map, Gaussian splatting, world model, cross-view geo-localization, traffic forecasting). We include works that propose or evaluate a representation for urban perception, prediction, planning, or simulation, or release urban data or protocols, and exclude work outside road mobility; per line of work we cite the first or most adopted method and its current representative. All entries were checked against official records; the supplementary material details the protocol.

\begin{figure*}[t]
\centering
\resizebox{\textwidth}{!}{%
\begin{tikzpicture}[
  font=\sffamily\scriptsize,
  head/.style={rounded corners=2pt, fill=wacvblue, text=white, font=\sffamily\bfseries\scriptsize, text width=3.1cm, minimum height=0.52cm, align=center, inner sep=1.5pt},
  item/.style={rounded corners=2pt, draw=wacvblue!60, fill=wacvblue!6, text width=3.0cm, minimum height=0.44cm, align=center, inner sep=1.5pt},
  coreitem/.style={rounded corners=2pt, draw=orange!60!black, fill=white, text width=2.75cm, minimum height=0.4cm, align=center, inner sep=1.5pt},
  arr/.style={-{Latex[length=2mm]}, thick, gray!70}
]
\def\xa{0} \def\xb{3.9} \def\xc{7.8} \def\xd{11.7} \def\xe{15.6}
\node[head] (h1) at (\xa,0) {Platforms $p$ (\S\ref*{sec:modalities})};
\node[head] (h2) at (\xb,0) {Write: $\Phi_p$, $\mathcal{U}$ (\S\ref*{sec:architectures})};
\node[head, fill=orange!80!black] (h3) at (\xc,0) {State $\mathcal{S}_t$ (\S\ref*{sec:representations})};
\node[head] (h4) at (\xd,0) {Read: $g_k$, $f$ (\S\ref*{sec:tasks}, \S\ref*{sec:applications})};
\node[head] (h5) at (\xe,0) {Evaluate (\S\ref*{sec:datasets}, \S\ref*{sec:usr5})};
\foreach \u/\v in {h1/h2,h2/h3,h3/h4,h4/h5} \draw[arr] (\u.east) -- (\v.west);
\foreach \t [count=\i] in {Ego: cam. / LiDAR / radar,Ego: thermal / event,Infra.: roadside C + L,V2V / V2I / V2X links,Aerial: drone / satellite,Traces: GNSS / loops,Priors: SD / HD map{,} POI}
  \node[item] at (\xa,-0.62*\i) {\t};
\foreach \t [count=\i] in {Early / mid / late fusion,View transforms,Cross-agent sharing,Temporal memory,Pretrained backbones,Modular vs.\ end-to-end,VLM / VLA heads}
  \node[item] at (\xb,-0.62*\i) {\t};
\draw[rounded corners=3pt, draw=orange!80!black, very thick, fill=orange!8] (\xc-1.6,-0.32) rectangle (\xc+1.6,-4.62);
\node[font=\sffamily\bfseries\scriptsize, orange!70!black] at (\xc,-0.56) {Urban State $\mathcal{S}_t$};
\foreach \t [count=\i] in {BEV feature grids,3D occupancy / voxels,Vectorized \& HD maps,Neural fields / 3DGS,Scene \& road graphs,Language tokens,World-model latents}
  \node[coreitem] at (\xc,-0.5-0.55*\i) {\t};
\foreach \t [count=\i] in {Detection \& segmentation,Tracking \& map upkeep,Motion forecasting,Planning \& control,Simulation \& twins,Traffic \& city analytics,QA / explanation}
  \node[item] at (\xd,-0.62*\i) {\t};
\foreach \t [count=\i] in {26 datasets (Tab.~\ref*{tab:datasets}),Perception metrics,Forecasting metrics,Open-loop planning,Closed-loop / pseudo-sim,Generative fidelity,Transfer \& robustness}
  \node[item] at (\xe,-0.62*\i) {\t};
\node[font=\sffamily\scriptsize\itshape, gray!40!black] at ($(\xb,-4.85)!0.5!(\xc,-4.85)$) {write $\Phi_p,\ \mathcal{U}$};
\node[font=\sffamily\scriptsize\itshape, gray!40!black] at ($(\xc,-4.85)!0.5!(\xd,-4.85)$) {read $g_k,\ f$};
\draw[rounded corners=2pt, dashed, gray] (-1.65,-5.1) rectangle (17.25,-5.55);
\node[font=\sffamily\scriptsize\itshape, gray!30!black] at (7.8,-5.325) {Cross-cutting (\S\ref*{sec:challenges}): cross-city \& cross-sensor transfer\,\textbullet\,uncertainty\,\textbullet\,efficiency \& bandwidth\,\textbullet\,open world, long tail \& drift\,\textbullet\,sim-to-real\,\textbullet\,privacy, safety \& governance};
\end{tikzpicture}}
\caption{\textbf{Taxonomy of Urban State Representation (\usr)}, organized by the terms of \cref{eq:usr}. Heterogeneous platforms \emph{write} to a shared, georeferenced, layered state $\mathcal{S}_t$ through encoders and fusion operators. Downstream tasks and applications \emph{read} from it, and datasets and protocols evaluate the whole chain. The bottom band lists challenges that cut across all stages.}
\label{fig:taxonomy}
\end{figure*}
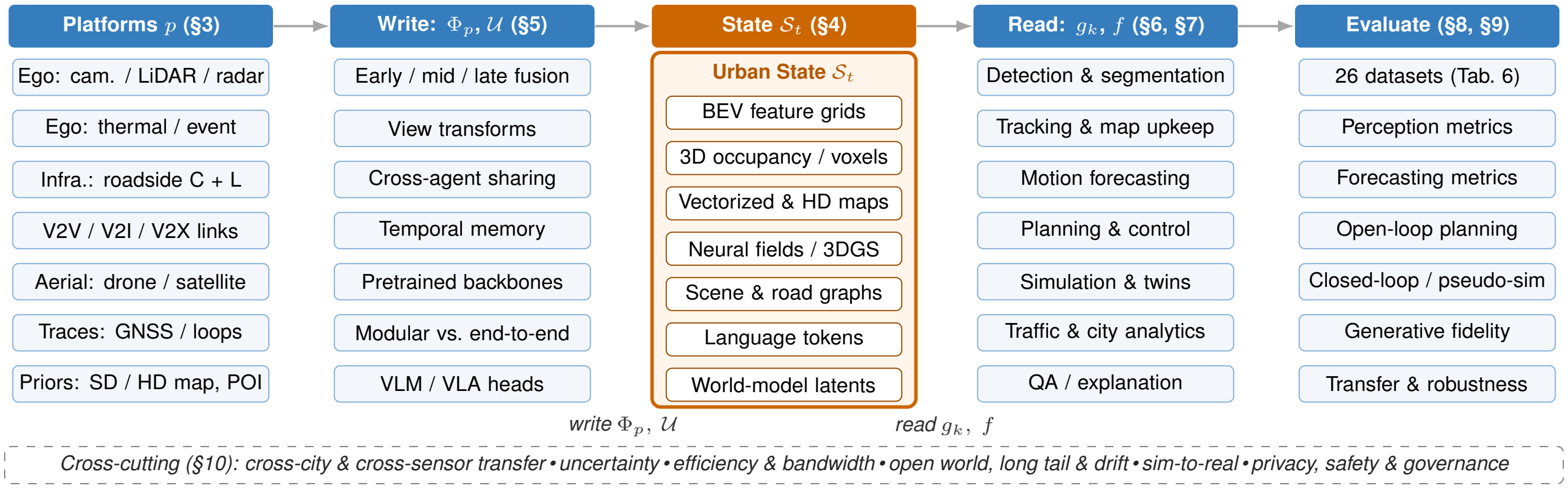

\noindent\textbf{Contributions.}
\textbf{(1)}~A formal, layered definition of the \emph{Urban State Representation} (\usr) with five desiderata (\cref{sec:formulation}).
\textbf{(2)}~A unified taxonomy (\cref{fig:taxonomy}) linking sensing platforms (\cref{sec:modalities}), eight representation families (\cref{sec:representations}), architectures (\cref{sec:architectures}), downstream tasks (\cref{sec:tasks}), and applications (\cref{sec:applications}).
\textbf{(3)}~A consolidated catalogue of 26 datasets and simulators, a metric glossary, a results snapshot across five benchmarks, a trend analysis, and a platform-coverage gap analysis (\cref{sec:datasets}).
\textbf{(4)}~\emph{USR-5}, a benchmark protocol of five tests with an aggregate score (\cref{sec:usr5}), and a research agenda pairing challenges with six open problems (\cref{sec:challenges}).
\noindent\textbf{Roadmap.} \Cref{fig:taxonomy} follows \cref{eq:usr} from left to right, and Secs.~3--8 follow its terms: platforms $p$ and observations $\mathbf{o}^p_t$ (\cref{sec:modalities}); the state $\mathcal{S}_t$, scored against D1--D5 (\cref{sec:representations}); the encoders $\Phi_p$ and update $\mathcal{U}$ that write it (\cref{sec:architectures}); the readouts $g_k$ and transition $f$ (\cref{sec:tasks,sec:applications}); and evaluation (\cref{sec:datasets,sec:usr5}). We present the state before its writers because architectures are best compared by the state they produce; the figure follows the data flow.

\begin{table}[t]
\centering
\caption{\textbf{Coverage of related surveys.} Plat.: multiple sensing platforms (V2X); Aer./Tr.: aerial imagery or mobility traces; Rep.: comparison across representation families; Arch.: architectures; WM: world models; FM: foundation or vision-language models; Data/Eval: dataset and evaluation-protocol analysis. \cmark~covered, \pmark~partial, \xmark~not covered. We trade depth on architectures, world models, and foundation models, which dedicated surveys cover, for breadth across platforms.}
\label{tab:surveys}
\scriptsize
\setlength{\tabcolsep}{2.4pt}
\begin{tabular}{@{}l c c c c c c c c c@{}}
\toprule
Survey & Year & Plat. & Aer./Tr. & Rep. & Arch. & WM & FM & Data & Eval\\
\midrule
BEV perception~\cite{li2024bevsurvey}          & 2024 & \xmark & \xmark & \pmark & \cmark & \xmark & \xmark & \cmark & \cmark\\
Occupancy~\cite{xu2025occsurvey}               & 2025 & \xmark & \xmark & \pmark & \cmark & \pmark & \xmark & \cmark & \cmark\\
Collab.\ perception~\cite{han2023collabsurvey} & 2023 & \cmark & \xmark & \xmark & \cmark & \xmark & \xmark & \cmark & \pmark\\
V2X perception~\cite{huang2025v2xcp}           & 2025 & \cmark & \xmark & \xmark & \cmark & \xmark & \pmark & \cmark & \pmark\\
End-to-end AD~\cite{chen2024e2esurvey}         & 2024 & \xmark & \xmark & \pmark & \cmark & \pmark & \pmark & \cmark & \cmark\\
MLLMs for AD~\cite{cui2024mllmsurvey}          & 2024 & \xmark & \xmark & \xmark & \pmark & \xmark & \cmark & \cmark & \pmark\\
3D Gaussians~\cite{chen2024gssurvey}           & 2024 & \xmark & \pmark & \pmark & \cmark & \xmark & \xmark & \cmark & \pmark\\
World models~\cite{feng2025wmsurvey}           & 2025 & \xmark & \xmark & \pmark & \cmark & \cmark & \pmark & \cmark & \pmark\\
Driving WMs~\cite{tu2026dwmsurvey}             & 2026 & \xmark & \xmark & \pmark & \cmark & \cmark & \pmark & \cmark & \cmark\\
Urban FMs~\cite{zhang2024ufmsurvey}            & 2024 & \pmark & \cmark & \xmark & \pmark & \xmark & \cmark & \cmark & \xmark\\
Shared WMs~\cite{hu2026maead}                   & 2026 & \cmark & \xmark & \pmark & \cmark & \cmark & \pmark & \cmark & \pmark\\
\midrule
\textbf{Ours}                                  & 2026 & \cmark & \cmark & \cmark & \pmark & \pmark & \pmark & \cmark & \cmark\\
\bottomrule
\end{tabular}
\end{table}

\section{Problem Formulation: the Urban State}
\label{sec:formulation}

Let $\mathcal{P}=\{p_1,\dots,p_N\}$ be the set of sensing platforms active in a region $\Omega\subset\mathbb{R}^3$. Platforms include ego vehicles, roadside units, aerial sensors, and trace providers. Each platform $p$ produces observations $\mathbf{o}^p_t$ of modality $m(p)$ in its own frame, with pose $\mathbf{T}^p_t\in SE(3)$ (or a 2D geolocation for traces), latency $\delta^p$, and bandwidth budget $b^p$. We define the \emph{Urban State Representation} as a latent field over space and time,
\begin{equation}
\mathcal{S}_t = \mathcal{U}\Big(\mathcal{S}_{t-1},\;\big\{\Phi_p\big(\mathbf{o}^p_{t-\delta^p},\mathbf{T}^p_{t-\delta^p}\big)\big\}_{p\in\mathcal{P}_t}\Big),
\label{eq:usr}
\end{equation}
where $\Phi_p$ is a platform-specific encoder that maps observations into a shared georeferenced frame and $\mathcal{U}$ is an update (fusion and temporal) operator. A task $k$ reads a prediction $\hat{y}^k_t = g_k(\mathcal{S}_t)$ and, ideally, an uncertainty $\hat{\Sigma}^k_t$. Planning and world models add an action-conditioned transition $\mathcal{S}_{t+1}\sim f(\mathcal{S}_t,\mathbf{a}_t)$.

\noindent\textbf{Special cases.} \Cref{eq:usr} specializes to familiar settings. With a single ego platform and $\mathcal{S}_t$ a BEV grid, it is BEV perception~\cite{li2022bevformer}. With several vehicles and intermediate feature sharing, it is collaborative perception~\cite{xu2022v2xvit}. With $\mathcal{U}$ an optimization over posed images, it is neural reconstruction~\cite{kerbl20233dgs}. With a learned generative $f$, it is a driving world model~\cite{hu2023gaia1}. With $\Phi_p$ aggregating loop-detector or map-matched GNSS readings and $\mathcal{S}_t$ a road-graph signal, it is traffic-state estimation~\cite{li2018dcrnn,newson2009mapmatching}.

\noindent\textbf{Role and benefits of USR.} USR is primarily a \emph{conceptual lens} that organizes existing work (Secs.~3--8), not a new method; the desiderata and USR-5 describe what a \emph{system} maintaining such a state would need, as targets rather than claims about existing systems. The unified view brings four benefits. (i)~\emph{Comparability}: works from separate communities are compared by which term of \cref{eq:usr} they change; cooperative perception and traffic-state estimation are both choices of $\Phi_p$ and $\mathcal{U}$. (ii)~\emph{Gap finding}: placing all platforms in one state exposes gaps no single community sees, \eg, no dataset pairs infrastructure with aerial views or traces (\cref{fig:coverage}). (iii)~\emph{Reuse}: a state written once and read by many heads amortizes computation (D2), and a persistent static layer need not be re-estimated each frame (D3). (iv)~\emph{Measurability}: it defines what ``shared'' means and so what to measure (\cref{sec:usr5}). Unlike digital twins, single-agent world models, or shared V2X world models~\cite{hu2026maead}, USR separates layers by update rate and admits aerial and trace writers.

\noindent\textbf{Layered state.} Urban content changes at very different rates, so we decompose the state into three layers,
\begin{equation}
\mathcal{S}_t = \big(\mathcal{S}^{\text{static}},\;\mathcal{S}^{\text{semi}}_{t},\;\mathcal{S}^{\text{dyn}}_{t}\big).
\label{eq:layers}
\end{equation}
The \emph{static} layer (road geometry, lane topology, buildings) changes over months and is best estimated by aggregating trips, aerial imagery, and maps~\cite{liao2023maptr,tancik2022blocknerf}. The \emph{semi-static} layer (parked vehicles, work zones, signal plans, congestion) changes over minutes to days and is observed well by traces and infrastructure~\cite{li2018dcrnn}. The \emph{dynamic} layer (moving agents, signal states) changes within seconds and needs low-latency on-board and V2X sensing~\cite{caesar2020nuscenes,yu2022dairv2x}. Layering makes explicit which platform should write which layer, and at what rate. Each layer favors different representations (\cref{tab:representations}, column L), and layers interact both ways: the static layer gives priors and a frame for dynamic observations, as in map-conditioned forecasting and planning~\cite{shi2022mtr,jiang2023vad}, and accumulated dynamic observations update slower layers, as in map-change detection~\cite{lambert2021tbv}.

\noindent\textbf{Integrating heterogeneous platforms.} The shared frame is map-anchored (ENU or UTM), not ego-centric. Integration follows \cref{eq:usr} in four steps: (1)~\emph{register} observations into it with $\mathbf{T}^p_t$, using cross-view matching for aerial images~\cite{zhu2021vigor} and map matching for traces~\cite{newson2009mapmatching}; (2)~\emph{synchronize} by compensating latency $\delta^p$, since latency and pose errors become feature misalignment, a major failure mode of cooperative fusion~\cite{xu2022v2xvit,huang2025v2xcp}; (3)~\emph{route} each platform to the layer matching its rate; and (4)~\emph{fuse} with $\mathcal{U}$, weighting by confidence and staleness within $b^p$~\cite{hu2022where2comm}. Steps (1)--(2) are mature in V2X perception; (3)--(4) across aerial and trace platforms remain open.

\noindent\textbf{Desiderata.} A useful \usr should be:
\textbf{(D1) platform-agnostic on write}, so that any $\Phi_p$ maps into the same space;
\textbf{(D2) task-agnostic on read}, so that many heads $g_k$ share $\mathcal{S}_t$;
\textbf{(D3) georeferenced and persistent}, so that knowledge accumulates across trips and agents rather than being rebuilt each frame;
\textbf{(D4) uncertainty-aware}, separating aleatoric and epistemic uncertainty~\cite{kendall2017uncertainties} and accounting for latency and pose error; and
\textbf{(D5) scalable}, in memory, compute, and communication, from an intersection to a city.

\section{Sensing Modalities and Platforms}
\label{sec:modalities}

\Cref{tab:modalities} summarizes the trade-offs. The platforms are complementary along three axes: \emph{viewpoint} (ego, elevated, overhead), \emph{spatial and temporal density}, and \emph{latency}. A \usr must therefore fuse observations that differ by orders of magnitude in resolution and update rate.

\noindent\textbf{Ego-vehicle sensors.} Cameras give dense semantics at low cost but no direct depth, and degrade at night and in bad weather~\cite{sakaridis2021acdc}; surround rigs of six to eight cameras are standard~\cite{caesar2020nuscenes,wilson2021argoverse2}. LiDAR gives metric geometry and underpins strong detectors~\cite{yin2021centerpoint}, but is sparse at range and costly. Radar measures radial velocity and is weather-robust; 4D imaging radar adds elevation and density~\cite{paek2022kradar}. Thermal and event cameras remain rare in public urban data.

\noindent\textbf{Infrastructure sensors.} Roadside cameras and LiDARs observe intersections from elevated, fixed, pre-calibrated viewpoints, reducing occlusion and giving long dwell times~\cite{yu2022dairv2x,yu2023v2xseq,zimmer2024tumtraf}. They are ideal writers of the semi-static layer and a persistent memory that passing vehicles can read.

\noindent\textbf{Aerial and satellite platforms.} Drones give occlusion-free, wide-area views for accurate trajectory extraction~\cite{krajewski2018highd} and detection and tracking at scale~\cite{zhu2022visdrone}; paired aerial and street views support city-scale reconstruction~\cite{li2023matrixcity}. Satellites provide global but infrequent coverage suited to the static layer. \emph{Cross-view matching} links the two viewpoints directly. Street-level images are retrieved against aerial tiles~\cite{workman2015cvusa,shi2019safa}, including off-center queries~\cite{zhu2021vigor} and transformer retrieval~\cite{zhu2022transgeo}, and vehicles can be localized metrically against satellite imagery, beyond retrieval~\cite{shi2022beyond}. They are the most mature bridge between ego and aerial platforms, and a ready-made writer of georeferenced poses $\mathbf{T}^p$.

\noindent\textbf{Mobility traces and priors.} GNSS traces are sparse in space but dense in time and city-wide~\cite{zheng2010geolife}; loop detectors give network speeds and flows~\cite{li2018dcrnn}; maps and points of interest encode slow structure. Traces are map-matched to the road graph, classically with hidden Markov models~\cite{newson2009mapmatching}; satellite imagery supports road extraction~\cite{demir2018deepglobe} and self-supervised pretraining~\cite{cong2022satmae}. These sources underpin urban computing~\cite{zheng2014urbancomputing,zhang2024ufmsurvey} but are rarely fused with sensor-level perception.

\begin{table}[t]
\centering
\caption{\textbf{Sensing modalities and platforms.} Qualitative comparison: H/M/L = high/medium/low. Res.: spatial resolution; Rng.: effective range; Wx.: robustness to weather and illumination; Cov.: area coverage; Layer: \usr layer it best serves (S static, M semi-static, D dynamic; \cref{eq:layers}).}
\label{tab:modalities}
\scriptsize
\setlength{\tabcolsep}{2.0pt}
\begin{tabular}{@{}l c c c c c c l@{}}
\toprule
Modality / platform & Res. & Rng. & Wx. & Cov. & Cost & Layer & Primary role\\
\midrule
Camera (ego)           & H & M & L & L & L & D   & Semantics, appearance\\
LiDAR (ego)            & M & M & M & L & H & D,S & Metric geometry\\
Radar / 4D radar       & L & H & H & L & L & D   & Velocity, all-weather\\
Thermal / event        & M & M & H & L & M & D   & Night, high dynamics\\
Roadside C+L           & M & M & M & M & M & D,M & Occlusion-free junctions\\
Drone / aerial         & H & H & L & H & M & S,M & Trajectories, maps\\
Satellite              & L & H & L & H & L & S   & Layout, change\\
GNSS / probe traces    & L & H & H & H & L & M   & Flow, demand, routing\\
Loop / traffic sensors & L & L & H & M & L & M   & Network traffic state\\
SD/HD map, POI         & -- & -- & H & H & M & S  & Static priors\\
\bottomrule
\end{tabular}
\end{table}

\section{Representation Families}
\label{sec:representations}

We group urban representations into eight families, each instantiating $\mathcal{S}_t$ differently (\cref{tab:representations}).

\noindent\textbf{BEV feature grids.} BEV maps perspective features onto a ground-plane grid, either by predicting per-pixel depth distributions and splatting (``lift-splat'')~\cite{philion2020lss} or by BEV queries that cross-attend to images with temporal self-attention~\cite{li2022bevformer}. It is the de facto common currency for multi-sensor~\cite{liu2023bevfusion} and multi-agent fusion~\cite{xu2022v2xvit,hu2022where2comm}, because any posed platform can project into it (D1)~\cite{li2024bevsurvey}. Its collapsed height axis, fixed ego-centric extent, and short temporal memory limit D3 and D5.

\noindent\textbf{3D occupancy.} Occupancy keeps the vertical axis and models ``un-boxable'' obstacles. Occ3D~\cite{tian2023occ3d} established large-scale semantic occupancy labels with visibility masks, and SurroundOcc~\cite{wei2023surroundocc} showed dense prediction from cameras alone~\cite{xu2025occsurvey}. Occupancy is a natural world-model state~\cite{zheng2024occworld} and fuses easily across platforms, but its cubic memory motivates sparse and Gaussian variants; GaussianFormer~\cite{huang2024gaussianformer} represents occupancy with semantic 3D Gaussians that adapt to scene content.

\noindent\textbf{Vectorized and HD maps.} Polylines and polygons encode lanes, boundaries, and crossings compactly. Online mapping predicts them from sensors~\cite{liao2023maptr}, and vectorized planners consume them~\cite{jiang2023vad}. Vectors are bandwidth-efficient (D5), shareable (D1), and persistent (D3), making them prime candidates for crowd-sourced maintenance of the static layer.

\noindent\textbf{Neural fields and 3D Gaussians.} NeRF~\cite{mildenhall2020nerf} represents scenes as radiance fields, and Block-NeRF~\cite{tancik2022blocknerf} scales this to city blocks. 3D Gaussian Splatting~\cite{kerbl20233dgs} rasterizes anisotropic Gaussians in real time and decomposes street scenes into background and moving actors~\cite{yan2024streetgaussians,chen2024gssurvey}, underpinning sensor simulation~\cite{yang2023unisim} and digital twins. These excel at D3 but are usually fit per scene offline; feed-forward reconstruction such as STORM~\cite{yang2025storm} predicts dynamic street Gaussians in a single pass, a step toward online use.

\noindent\textbf{Scene graphs.} Nodes (agents, lanes, signals) and edges (relations, right-of-way, interactions) encode the relational structure that planning and explanation need. Graph-structured QA over driving scenes~\cite{sima2024drivelm} shows how graphs can bridge perception, reasoning, and language. Relational context also aids open-world perception (\cref{sec:tasks}). Graphs are compact and task-agnostic, but rarely carry calibrated uncertainty.

\noindent\textbf{Road-network graphs.} Urban computing represents the city as signals on the nodes and edges of the road graph, such as speed, flow, or demand. Spatio-temporal graph networks forecast these signals with diffusion convolutions~\cite{li2018dcrnn}, graph convolutions combined with temporal gating~\cite{yu2018stgcn}, or adaptive, learned adjacency~\cite{wu2019graphwavenet}. Road graphs are compact (D5), persistent (D3), and natural writers of the semi-static layer. However, they have no lane-level geometry, so they are rarely connected to AV perception. Anchoring road-graph signals to vectorized lane maps~\cite{liao2023maptr} is a direct way to join the two.

\noindent\textbf{Language and token representations.} Vision-language models compress scenes into text or tokens that can be reasoned over and explained~\cite{tian2024drivevlm,hwang2024emma,cui2024mllmsurvey}, and image-text pretraining~\cite{radford2021clip} supplies open-vocabulary semantics. Language maximizes D2 and interpretability at the cost of metric precision and latency.

\noindent\textbf{World-model latents.} Generative world models learn a latent state and its dynamics, predicting future video~\cite{hu2023gaia1} or occupancy~\cite{zheng2024occworld} under actions~\cite{feng2025wmsurvey,tu2026dwmsurvey}. They instantiate the predictive part of \cref{eq:usr} but are rarely georeferenced, persistent, or shared across agents.

\noindent\textbf{Hybrid states.} No family meets all five desiderata (\cref{tab:representations}). Current systems combine families implicitly: UniAD~\cite{hu2023uniad} chains BEV, track, map, and occupancy queries; VAD~\cite{jiang2023vad} pairs BEV with vectors; DriveVLM~\cite{tian2024drivevlm} couples language with a metric planner. We expect explicit hybrids in which vectors and road graphs hold topology, occupancy or Gaussians hold geometry and appearance, tracks hold the dynamic layer, and tokens hold semantics, all bound to one georeferenced frame.

\noindent\textbf{Trade-offs.} Three trade-offs explain the scorecard. \emph{Density vs.\ bandwidth}: sharing dense intermediate features beats sharing boxes on DAIR-V2X (63.7 vs.\ 53.1 AP@0.5, \cref{tab:results}) but costs bandwidth, which Where2comm~\cite{hu2022where2comm} cuts by sending only confident regions; vectors and graphs are compact but drop appearance. \emph{Persistence vs.\ freshness}: persistent maps and fields amortize estimation but go stale after change~\cite{lambert2021tbv}. \emph{Explicit vs.\ latent}: explicit states can be verified and calibrated; latents and tokens enable generation and reasoning but are hard to georeference or audit.

\begin{table}[t]
\centering
\caption{\textbf{Representation families versus \usr desiderata} (\cref{sec:formulation}). \emph{Rubric.} D1~\cmark: any posed platform can write without retraining the state. D2~\cmark: three or more task heads shown on one state. D3~\cmark: map-anchored and reused across traversals. D4~\cmark: per-element uncertainty is output. D5~\cmark: memory or bandwidth grows sub-linearly with area, or the state is sparse or vectorized. \pmark: met by some published variants; \xmark: not met in typical practice. L: layer it mainly holds (St static, Se semi-static, Dy dynamic). Marks are judgments under this rubric, justified per mark in the supplementary material.}
\label{tab:representations}
\scriptsize
\setlength{\tabcolsep}{2.6pt}
\begin{tabular}{@{}l c c c c c c l@{}}
\toprule
Family & D1 & D2 & D3 & D4 & D5 & L & Representative\\
\midrule
BEV grid            & \cmark & \cmark & \pmark & \pmark & \pmark & Dy & \cite{philion2020lss,li2022bevformer,liu2023bevfusion}\\
3D occupancy        & \cmark & \cmark & \pmark & \pmark & \xmark & Dy & \cite{tian2023occ3d,wei2023surroundocc}\\
Vectorized map      & \cmark & \pmark & \cmark & \pmark & \cmark & St & \cite{liao2023maptr,jiang2023vad}\\
Neural field / 3DGS & \pmark & \pmark & \cmark & \xmark & \pmark & St & \cite{tancik2022blocknerf,kerbl20233dgs,yan2024streetgaussians}\\
Scene graph         & \cmark & \cmark & \pmark & \xmark & \cmark & Dy & \cite{sima2024drivelm}\\
Language tokens     & \pmark & \cmark & \xmark & \xmark & \cmark & Dy & \cite{tian2024drivevlm,hwang2024emma}\\
World-model latent  & \xmark & \pmark & \xmark & \pmark & \pmark & Dy & \cite{hu2023gaia1,zheng2024occworld}\\
Road-network graph  & \pmark & \pmark & \cmark & \pmark & \cmark & Se & \cite{li2018dcrnn,yu2018stgcn,wu2019graphwavenet}\\
\bottomrule
\end{tabular}
\end{table}

\section{Architectures}
\label{sec:architectures}

\noindent\textbf{Fusion stage.} \emph{Early} fusion combines raw data and preserves information but is sensitive to calibration and bandwidth; \emph{late} fusion combines detections cheaply but discards evidence; \emph{intermediate} fusion dominates, \eg, BEVFusion~\cite{liu2023bevfusion} unifies camera and LiDAR in BEV and TransFuser~\cite{chitta2023transfuser} fuses them by attention.

\noindent\textbf{Cross-platform fusion.} V2X-ViT~\cite{xu2022v2xvit} handles heterogeneous agents with latency- and pose-aware attention; Where2comm~\cite{hu2022where2comm} transmits only informative features chosen by confidence maps; UniV2X~\cite{yu2025univ2x} extends cooperation to end-to-end planning. Asynchrony, localization error, and model heterogeneity~\cite{han2023collabsurvey,huang2025v2xcp} are exactly the $\delta^p$ and $\mathbf{T}^p$ terms of \cref{eq:usr}. Aerial imagery and traces, the natural writers of the static and semi-static layers, remain absent.

\noindent\textbf{Temporal modeling.} Recurrent BEV states~\cite{li2022bevformer}, object-centric temporal queries~\cite{wang2023streampetr}, and track queries~\cite{hu2023uniad} propagate state over time, and world models make this operator generative~\cite{zheng2024occworld}. How long memory should last, seconds for agents or days for context, is open.

\noindent\textbf{3D geometric foundation models.} New models regress 3D structure from uncalibrated images: DUSt3R~\cite{wang2024dust3r} predicts dense point maps from image pairs, MASt3R~\cite{leroy2024mast3r} adds 3D-grounded matching, and VGGT~\cite{wang2025vggt} infers cameras, depth, and point maps for many views in one pass; monocular depth models~\cite{yang2024depthanything} transfer across cameras. They approach the platform-agnostic $\Phi_p$ of D1, placing dashcam, roadside, and drone imagery in one frame without per-rig calibration, though metric scale and latency on driving data remain open.

\noindent\textbf{Backbones and pretraining.} Vision foundation models such as CLIP~\cite{radford2021clip} and SAM~\cite{kirillov2023sam} supply transferable semantics, while urban and satellite foundation models~\cite{zhang2024ufmsurvey,cong2022satmae} remain disconnected from driving backbones; bridging them is a central opportunity for \usr.

\noindent\textbf{Modular versus end-to-end.} Classical stacks chain detection~\cite{yin2021centerpoint}, tracking, mapping~\cite{liao2023maptr}, forecasting~\cite{shi2022mtr}, and planning. End-to-end models optimize the chain jointly~\cite{chen2024e2esurvey}: UniAD~\cite{hu2023uniad} links tracking, mapping, occupancy, and motion queries to planning; VAD~\cite{jiang2023vad} plans over vectorized tokens; TransFuser~\cite{chitta2023transfuser} maps fused sensors to waypoints; and diffusion-based planners such as DiffusionDrive~\cite{liao2025diffusiondrive} generate multimodal trajectories in a few denoising steps, achieving state-of-the-art NAVSIM results at publication. End-to-end training makes the \usr implicit and task-optimized, helping D2 within one model but hurting interpretability and reuse across platforms.

\noindent\textbf{VLM and VLA models.} DriveVLM~\cite{tian2024drivevlm} pairs a VLM for description and hierarchical planning with a fast planner; EMMA~\cite{hwang2024emma} casts perception, road-graph estimation, and planning as text generation; AutoVLA~\cite{zhou2025autovla} adapts how much it reasons per scene and is refined by reinforcement fine-tuning. They bring world knowledge and long-tail reasoning but raise latency, grounding, and hallucination concerns~\cite{cui2024mllmsurvey}.

\noindent\textbf{World models and neural simulation.} Video~\cite{hu2023gaia1,wang2024drivedreamer,gao2024vista} and occupancy~\cite{zheng2024occworld} world models learn $f(\mathcal{S}_t,\mathbf{a}_t)$, and neural simulators~\cite{yang2023unisim} re-render real logs under counterfactual actions, turning the \usr from a perception product into a simulation substrate~\cite{feng2025wmsurvey,tu2026dwmsurvey}. \Cref{tab:paradigms} compares the paradigms.

\noindent\textbf{Generative data engines.} The same generative machinery can \emph{write} training data. MagicDrive~\cite{gao2024magicdrive} synthesizes multi-camera street views controlled by 3D boxes, road maps, and camera poses. DriveDreamer~\cite{wang2024drivedreamer} and Vista~\cite{gao2024vista} generate controllable future scenes, and world foundation models such as Cosmos~\cite{nvidia2025cosmos} are released as pretrained platforms for physical AI. Conditioned on a \usr, they could generate rare scenarios for the long tail, if validated (P5).

\begin{table}[t]
\centering
\caption{\textbf{Architectural paradigms} and the \usr they produce. \emph{Rubric.} Interp.\ H: the state is human-readable (objects, maps, text); M: spatially grounded features; L: opaque latents. X-plat.\ H: the state is metric and frame-aligned, so other platforms can fuse it; L: it is tied to one model. Lat.\ (relative inference latency) L: real-time on-vehicle; M: real-time with large backbones or V2X delay; H: offline or LLM-scale.}
\label{tab:paradigms}
\scriptsize
\setlength{\tabcolsep}{2.2pt}
\begin{tabular}{@{}l l c c c l@{}}
\toprule
Paradigm & Implied \usr & Interp. & X-plat. & Lat. & Examples\\
\midrule
Geometric FM       & Point maps / poses     & M & H & M & \cite{wang2024dust3r,wang2025vggt}\\
Modular stack      & Explicit objects + map & H & H & L & \cite{yin2021centerpoint,liao2023maptr,shi2022mtr}\\
Sensor fusion (BEV)& Shared BEV features    & M & H & L & \cite{liu2023bevfusion,li2022bevformer}\\
Cooperative fusion & Multi-agent BEV        & M & H & M & \cite{xu2022v2xvit,hu2022where2comm,yu2025univ2x}\\
End-to-end planner & Task-optimized queries & L & L & M & \cite{hu2023uniad,jiang2023vad,chitta2023transfuser}\\
VLM / VLA          & Language tokens        & H & M & H & \cite{tian2024drivevlm,hwang2024emma}\\
World model        & Predictive latent      & L & L & H & \cite{hu2023gaia1,zheng2024occworld}\\
Neural simulator   & Fields / Gaussians     & M & M & H & \cite{yang2023unisim,yan2024streetgaussians}\\
\bottomrule
\end{tabular}
\end{table}

\section{Downstream Tasks}
\label{sec:tasks}

\noindent\textbf{Perception.} 3D detection~\cite{yin2021centerpoint,li2022bevformer,liu2023bevfusion}, semantic segmentation~\cite{cordts2016cityscapes}, and occupancy prediction~\cite{tian2023occ3d,wei2023surroundocc} are the first readouts of the dynamic and static layers. Cooperative variants evaluate the same heads under communication constraints~\cite{xu2022opv2v,yu2022dairv2x,xu2023v2v4real}.

\noindent\textbf{Open-world and context-aware perception.} A \usr must record objects outside a detector's label set, because unknown obstacles concentrate risk. The \emph{geometry-first} route writes them into occupancy as occupied but unlabeled space~\cite{tian2023occ3d}; the \emph{semantics-first} route opens the label space with image-text features~\cite{radford2021clip} or unknown-aware contrastive learning over co-occurrence context~\cite{singh2026unknown}. Context also helps closed-set recognition: fusing region-level and scene-level context improves small, rare, and occluded objects on Cityscapes and BDD100K~\cite{singh2026ccff}. These object relations form the scene-graph layer of a \usr.

\noindent\textbf{Tracking and map maintenance.} Tracking and online vectorized mapping~\cite{liao2023maptr} maintain persistent entities and topology (D3); detecting map changes from vehicle data closes the read-write loop on the static layer~\cite{lambert2021tbv}.

\noindent\textbf{Motion forecasting.} Transformer forecasters~\cite{shi2022mtr} predict multimodal futures on large motion datasets~\cite{ettinger2021womd,wilson2021argoverse2}; infrastructure~\cite{yu2023v2xseq} and drone data~\cite{krajewski2018highd} add longer, less occluded histories.

\noindent\textbf{Planning.} Planners read the \usr to produce trajectories. They are evaluated open-loop on logs, in pseudo-simulation~\cite{dauner2024navsim}, or closed-loop in simulators~\cite{dosovitskiy2017carla,jia2024bench2drive,caesar2021nuplan}.

\noindent\textbf{Simulation and twins.} Reconstructive~\cite{yan2024streetgaussians,yang2023unisim} and generative~\cite{gao2024magicdrive,gao2024vista} representations and city-scale data~\cite{li2023matrixcity} support sensor simulation, scenario editing, and persistent twins.

\noindent\textbf{Traffic and city analytics.} Traffic forecasting on road graphs~\cite{li2018dcrnn,yu2018stgcn,wu2019graphwavenet} and mobility mining~\cite{zheng2010geolife} read the same physical state at coarser resolution, a natural but under-used readout of a shared \usr.

\noindent\textbf{Question answering and explanation.} Language readouts~\cite{sima2024drivelm,tian2024drivevlm} expose the \usr to humans for auditing and interaction.

\section{Applications}
\label{sec:applications}

\noindent\textbf{Fleets and infrastructure.} Robotaxi fleets repeatedly traverse the same streets, so each vehicle can read a prior \usr and write corrections, amortizing perception and keeping maps current~\cite{liao2023maptr,tancik2022blocknerf}. Roadside units at complex intersections share perception or planning-ready features with approaching vehicles to resolve occlusions~\cite{yu2022dairv2x,yu2025univ2x}.

\noindent\textbf{Traffic management and digital twins.} The semi-static layer built from loops, traces, and roadside sensors~\cite{li2018dcrnn,yu2018stgcn} supports signal timing and congestion forecasting, enriched by lane-level AV perception. Neural reconstructions~\cite{yan2024streetgaussians,yang2023unisim} and world models~\cite{hu2023gaia1} let agencies replay incidents and test designs before deployment.

\noindent\textbf{Layer assignment.} At an instrumented intersection, fleet maps and aerial surveys write the static layer; roadside units, loops, and traces the semi-static layer; and AVs and roadside units the dynamic layer at 10--20\,Hz.

\noindent\textbf{Robust operations.} Adverse-weather sensing~\cite{paek2022kradar,sakaridis2021acdc} and redundant infrastructure and aerial views keep the state usable when one vehicle's sensors fail.

\begin{table*}[t]
\centering
\caption{\textbf{Research agenda for Urban State Representation.} Each challenge is paired with an open problem, promising directions from the surveyed literature, and the quantities a benchmark should report.}
\label{tab:agenda}
\scriptsize
\setlength{\tabcolsep}{3pt}
\begin{tabularx}{\textwidth}{@{}l l X l@{}}
\toprule
Challenge & Open problem & Promising directions & What to measure\\
\midrule
Data gaps & (P1) Joint multi-platform datasets & Co-register V2I sets~\cite{yu2022dairv2x} with drone, satellite~\cite{zhu2021vigor} and traffic feeds~\cite{li2018dcrnn} & Platform coverage (\cref{fig:coverage}); \#cities\\
Transfer & (P2) Write-anywhere encoders & Geometric FMs~\cite{wang2025vggt}; foundation priors~\cite{radford2021clip,kirillov2023sam} & Drop on held-out cities / rigs~\cite{wang2020traingermany}\\
Efficiency & (P3) Persistent city-scale state & Vector maps~\cite{liao2023maptr}; streaming Gaussians~\cite{yan2024streetgaussians}; change detection~\cite{lambert2021tbv} & Accuracy vs.\ bytes~\cite{hu2022where2comm}; staleness\\
Uncertainty & (P4) Calibrated uncertainty end to end & Latency-aware fusion~\cite{xu2022v2xvit}; confidence-driven sharing~\cite{hu2022where2comm} & Calibration error; unknown recall\\
Sim-to-real & (P5) Trustworthy generative evaluation & Neural sim.~\cite{yang2023unisim}; generative engines~\cite{gao2024magicdrive,gao2024vista}; pseudo-sim.~\cite{dauner2024navsim} & Sim--real rank correlation of planners\\
Community silos & (P6) Bridging AV and urban computing & Link driving and urban FMs~\cite{zhang2024ufmsurvey} & AV + traffic readouts~\cite{li2018dcrnn}\\
\bottomrule
\end{tabularx}
\end{table*}

\section{Datasets, Benchmarks, and Metrics}
\label{sec:datasets}

\noindent\textbf{Datasets.} \Cref{tab:datasets} consolidates 26 datasets and simulators in six settings: ego-centric~\cite{geiger2012kitti,cordts2016cityscapes,caesar2020nuscenes,sun2020waymo,yu2020bdd100k,wilson2021argoverse2}, adverse conditions~\cite{sakaridis2021acdc,paek2022kradar}, cooperative V2V/V2I, simulated~\cite{xu2022opv2v} and real~\cite{yu2022dairv2x,xu2023v2v4real,yu2023v2xseq,zimmer2024tumtraf}, aerial and satellite~\cite{krajewski2018highd,zhu2022visdrone,demir2018deepglobe,li2023matrixcity}, traces and traffic~\cite{zheng2010geolife,li2018dcrnn}, and planning and language~\cite{caesar2021nuplan,dauner2024navsim,jia2024bench2drive,sima2024drivelm}. Derived benchmarks re-annotate existing logs, \eg, Occ3D~\cite{tian2023occ3d} adds occupancy labels to nuScenes and Waymo.

\begin{table*}[t]
\centering
\caption{\textbf{Urban datasets and simulators for \usr research}, grouped by platform setting. C: camera; L: LiDAR; R: radar; 4DR: 4D radar; G: GNSS trajectories; Lp: loop detectors; Q: language QA. Scale figures are approximate and follow each original paper; ``sim'' marks synthetic data.}
\label{tab:datasets}
\scriptsize
\setlength{\tabcolsep}{3.2pt}
\begin{tabularx}{\textwidth}{@{}l l c l l l X@{}}
\toprule
Setting & Dataset & Year & Modalities & Scale & Region & Primary tasks\\
\midrule
\multirow{8}{*}{Ego-centric}
 & KITTI~\cite{geiger2012kitti}               & 2012 & C, L, IMU    & $\sim$15k annotated frames & Karlsruhe (DE) & 3D detection, tracking, stereo, odometry\\
 & Cityscapes~\cite{cordts2016cityscapes}     & 2016 & C (stereo)   & 5k fine + 20k coarse images & 50 cities (EU) & Semantic / instance segmentation\\
 & nuScenes~\cite{caesar2020nuscenes}         & 2020 & 6C, 1L, 5R   & 1k scenes, 40k keyframes & Boston, Singapore & Detection, tracking, prediction, segmentation\\
 & Waymo Open~\cite{sun2020waymo}             & 2020 & 5C, 5L       & 1,150 scenes & 3 US cities & 2D/3D detection, tracking\\
 & BDD100K~\cite{yu2020bdd100k}              & 2020 & C, G (dashcam) & 100k videos & NYC, SF Bay Area (US) & Detection, segmentation, lanes, drivable area, tracking\\
 & WOMD~\cite{ettinger2021womd}               & 2021 & Boxes, map   & $>$100k 20\,s scenes & 6 US cities & Interactive motion forecasting\\
 & Argoverse 2~\cite{wilson2021argoverse2}    & 2021 & C, L, map    & 1k sensor logs, 250k motion scenarios & 6 US cities & Detection, tracking, forecasting\\
 & Occ3D~\cite{tian2023occ3d}                 & 2023 & C, L (derived) & nuScenes + Waymo & Multi-city & Semantic 3D occupancy\\
\midrule
\multirow{2}{*}{Conditions}
 & ACDC~\cite{sakaridis2021acdc}              & 2021 & C            & 4,006 images & Switzerland & Segmentation in fog, night, rain, snow\\
 & K-Radar~\cite{paek2022kradar}              & 2022 & 4DR, C, L    & 35k frames & South Korea & All-weather 3D detection\\
\midrule
\multirow{5}{*}{Cooperative}
 & OPV2V~\cite{xu2022opv2v}                   & 2022 & C, L (sim)   & 11.5k frames, $>$70 scenes & CARLA & V2V cooperative detection\\
 & DAIR-V2X~\cite{yu2022dairv2x}              & 2022 & C, L         & 71k frames & Beijing (CN) & V2I cooperative 3D detection\\
 & V2V4Real~\cite{xu2023v2v4real}             & 2023 & C, L         & 20k LiDAR frames & Ohio (US) & Real V2V detection, tracking, sim2real\\
 & V2X-Seq~\cite{yu2023v2xseq}                & 2023 & C, L, map    & 15k frames + 210k scenarios & Beijing (CN) & V2I tracking and forecasting\\
 & TUMTraf V2X~\cite{zimmer2024tumtraf}       & 2024 & C, L         & 2k point clouds, 5k images & Munich (DE) & V2I detection and tracking\\
\midrule
\multirow{4}{*}{\shortstack[l]{Aerial \&\\satellite}}
 & highD~\cite{krajewski2018highd}            & 2018 & C (drone)    & 110k vehicles, 16.5\,h & German highways & Naturalistic trajectories\\
 & VisDrone~\cite{zhu2022visdrone}            & 2022 & C (drone)    & 263 clips + 10k images & 14 cities (CN) & Aerial detection and tracking\\
 & DeepGlobe~\cite{demir2018deepglobe}        & 2018 & C (satellite) & Road, building, land-cover tiles & Multi-country & Road extraction, land-cover segmentation\\
 & MatrixCity~\cite{li2023matrixcity}         & 2023 & C (sim)      & $\sim$519k aerial + street images & Synthetic city & City-scale neural rendering\\
\midrule
\multirow{2}{*}{Traces}
 & GeoLife~\cite{zheng2010geolife}            & 2010 & G            & 17.6k trajectories, 182 users & Beijing (CN) & Mobility mining, mode inference\\
 & METR-LA~\cite{li2018dcrnn}                 & 2018 & Lp           & 207 sensors, 4 months & Los Angeles (US) & Traffic-state forecasting\\
\midrule
\multirow{5}{*}{\shortstack[l]{Planning \&\\language}}
 & CARLA~\cite{dosovitskiy2017carla}          & 2017 & C, L, R (sim) & Simulator & Synthetic towns & Closed-loop driving\\
 & nuPlan~\cite{caesar2021nuplan}             & 2021 & C, L         & $\sim$1,500\,h driving & 4 cities & Closed-loop planning\\
 & NAVSIM~\cite{dauner2024navsim}             & 2024 & C, L         & nuPlan-derived & 4 cities & Non-reactive pseudo-simulation (PDMS)\\
 & Bench2Drive~\cite{jia2024bench2drive}      & 2024 & C, L, R (sim) & 2M frames, 220 routes & CARLA & Closed-loop end-to-end, 44 scenarios\\
 & DriveLM~\cite{sima2024drivelm}             & 2024 & C, Q         & nuScenes + CARLA & Multi & Graph VQA, explainable planning\\
\bottomrule
\end{tabularx}
\end{table*}

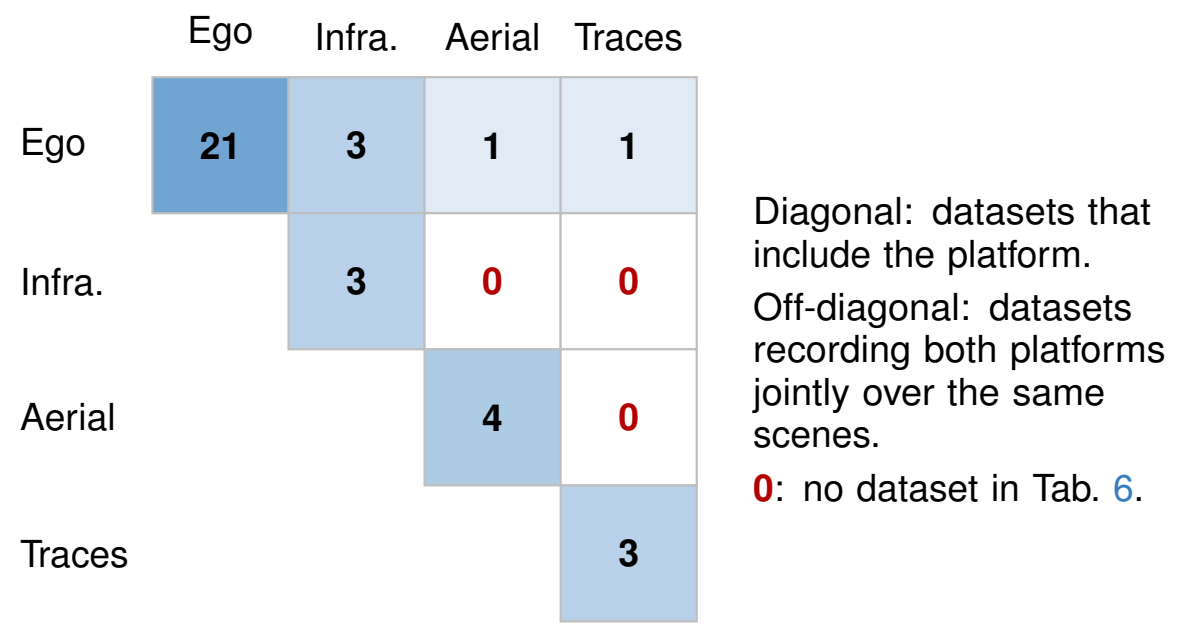
\begin{figure}[t]
\centering
\begin{tikzpicture}[font=\sffamily\scriptsize]
\def\cs{0.9}
\foreach \l [count=\i] in {Ego,Infra.,Aerial,Traces}{
  \node[anchor=west] at (-1.0,-\i*\cs+0.45) {\l};
  \node[anchor=south, rotate=0] at (\i*\cs-0.45,0.05) {\l};}
\foreach \i/\j/\v/\c in {1/1/21/70,1/2/3/35,1/3/1/15,1/4/1/15,2/2/3/35,2/3/0/0,2/4/0/0,3/3/4/40,3/4/0/0,4/4/3/35}{
  \fill[wacvblue!\c] (\j*\cs-\cs,-\i*\cs) rectangle (\j*\cs,-\i*\cs+\cs);
  \draw[gray!50] (\j*\cs-\cs,-\i*\cs) rectangle (\j*\cs,-\i*\cs+\cs);
  \node at (\j*\cs-0.45,-\i*\cs+0.45) {\ifnum\v=0 \textcolor{red!70!black}{\textbf{0}}\else\textbf{\v}\fi};}
\node[anchor=west, align=left, text width=3.1cm] at (4*\cs+0.25,-1.8) {Diagonal: datasets that include the platform.\\[2pt] Off-diagonal: datasets recording both platforms jointly over the same scenes.\\[2pt] \textcolor{red!70!black}{\textbf{0}}: no dataset in Tab.~\ref{tab:datasets}.};
\end{tikzpicture}
\caption{\textbf{Platform coverage across the 26 datasets in \cref{tab:datasets}.} Ego--infrastructure: DAIR-V2X, V2X-Seq, TUMTraf. Ego(street)--aerial: MatrixCity (synthetic only). Ego--traces: BDD100K, whose dashcam GPS logs describe the recording vehicles' own trips rather than city-wide mobility. Multi-vehicle (V2V) pairing appears in OPV2V and V2V4Real. No dataset here pairs infrastructure with aerial views or traces.}
\label{fig:coverage}
\end{figure}

\noindent\textbf{Platform-coverage gap.} \Cref{fig:coverage} counts datasets in \cref{tab:datasets} that record each pair of platforms jointly. Ego data dominate (21 of 26). Cooperative datasets cover ego--infrastructure and multi-vehicle pairs; aerial--street pairing is synthetic~\cite{li2023matrixcity}, or real but without driving sensors~\cite{zhu2021vigor}; and perception meets traces only via BDD100K's dashcam GPS~\cite{yu2020bdd100k}. Data also concentrate in a few cities in the US, Europe, and East Asia, limiting cross-city study~\cite{wang2020traingermany}.

\noindent\textbf{Trends.} \Cref{tab:datasets} shows four shifts: single-task to multi-task, multi-sensor suites; one ego vehicle to cooperative recordings (yearly since 2022); open-loop logs to closed-loop and pseudo-simulation; and labels to language~\cite{sima2024drivelm}.

\noindent\textbf{Metrics.} \Cref{tab:metrics} summarizes metrics across readouts. Open-loop planning metrics can be dominated by ego-status shortcuts~\cite{li2024egostatus}, and cooperative accuracy should be reported with bandwidth and latency~\cite{hu2022where2comm}.

\begin{table}[t]
\centering
\caption{\textbf{Metric glossary} for \usr readouts.}
\label{tab:metrics}
\scriptsize
\setlength{\tabcolsep}{2.5pt}
\begin{tabularx}{\columnwidth}{@{}l l X@{}}
\toprule
Task & Metric & Definition / used in\\
\midrule
3D detection   & mAP, NDS       & Center-distance AP; NDS adds box-error terms~\cite{caesar2020nuscenes}\\
Segmentation   & mIoU           & Mean intersection-over-union~\cite{cordts2016cityscapes}\\
Occupancy      & mIoU (voxel)   & Per-voxel semantic IoU, visible voxels~\cite{tian2023occ3d}\\
Tracking       & AMOTA          & MOTA averaged over recall thresholds~\cite{caesar2020nuscenes}\\
Forecasting    & minADE/FDE, MR & Best-of-$K$ displacement, miss rate~\cite{ettinger2021womd,wilson2021argoverse2}\\
Open-loop plan & L2, collision  & Distance to log trajectory; shortcut-prone~\cite{li2024egostatus}\\
Pseudo-sim     & PDMS           & Collision, drivable area, progress, comfort~\cite{dauner2024navsim}\\
Closed-loop    & DS, success    & Route completion $\times$ infraction penalty~\cite{jia2024bench2drive}\\
Cooperative    & AP@IoU + bytes & Accuracy vs.\ bandwidth curve~\cite{hu2022where2comm}\\
Rendering      & PSNR/SSIM/LPIPS& Novel-view fidelity~\cite{kerbl20233dgs,yan2024streetgaussians}\\
Generation     & FID / FVD      & Image / video distribution distance~\cite{hu2023gaia1}\\
Traffic state  & MAE / RMSE     & Speed or flow error per horizon~\cite{li2018dcrnn}\\
Language       & Acc., BLEU     & QA accuracy and text overlap~\cite{sima2024drivelm}\\
\bottomrule
\end{tabularx}
\end{table}

\begin{figure}[t]
\centering
\begin{tikzpicture}
\begin{axis}[
  width=\columnwidth, height=5.0cm,
  xmin=2021.2, xmax=2024.05, ymin=40, ymax=78,
  xtick={2021.5,2022.5,2023.5}, xticklabels={2021,2022,2023},
  ylabel={nuScenes test NDS}, xlabel={Approximate publication date},
  tick label style={font=\scriptsize}, label style={font=\scriptsize},
  legend style={font=\tiny, at={(0.98,0.03)}, anchor=south east, draw=gray!40},
  grid=major, grid style={gray!18}, clip=false]
\addplot[dashed, wacvblue, mark=*, mark options={solid, fill=wacvblue}] coordinates {(2021.8,42.8) (2021.87,47.9) (2022.8,56.9) (2023.8,67.6)};
\addlegendentry{Camera}
\addplot[only marks, mark=square*, color=gray!60!black] coordinates {(2021.45,67.3) (2022.45,70.2)};
\addlegendentry{LiDAR}
\addplot[only marks, mark=triangle*, mark size=3pt, color=orange!85!black] coordinates {(2022.48,71.7) (2023.37,72.9)};
\addlegendentry{Camera + LiDAR}
\node[font=\tiny, anchor=west] at (axis cs:2021.83,42.0) {FCOS3D};
\node[font=\tiny, anchor=west] at (axis cs:2021.92,48.6) {DETR3D};
\node[font=\tiny, anchor=west] at (axis cs:2022.84,56.0) {BEVFormer};
\node[font=\tiny, anchor=east] at (axis cs:2023.76,67.6) {StreamPETR};
\node[font=\tiny, anchor=north] at (axis cs:2021.45,66.6) {CenterPoint};
\node[font=\tiny, anchor=north west] at (axis cs:2022.52,70.9) {TransFusion(-L)};
\node[font=\tiny, anchor=south] at (axis cs:2023.37,73.4) {BEVFusion};
\end{axis}
\end{tikzpicture}
\caption{\textbf{Progress on nuScenes 3D detection (test NDS) by sensing modality}, plotted at approximate publication dates, as reported in~\cite{wang2023streampetr,bai2022transfusion,yin2021centerpoint,liu2023bevfusion,li2022bevformer,wang2021detr3d}. Camera-only methods closed much of the gap to LiDAR, largely through temporal state modeling.}
\label{fig:nds}
\end{figure}
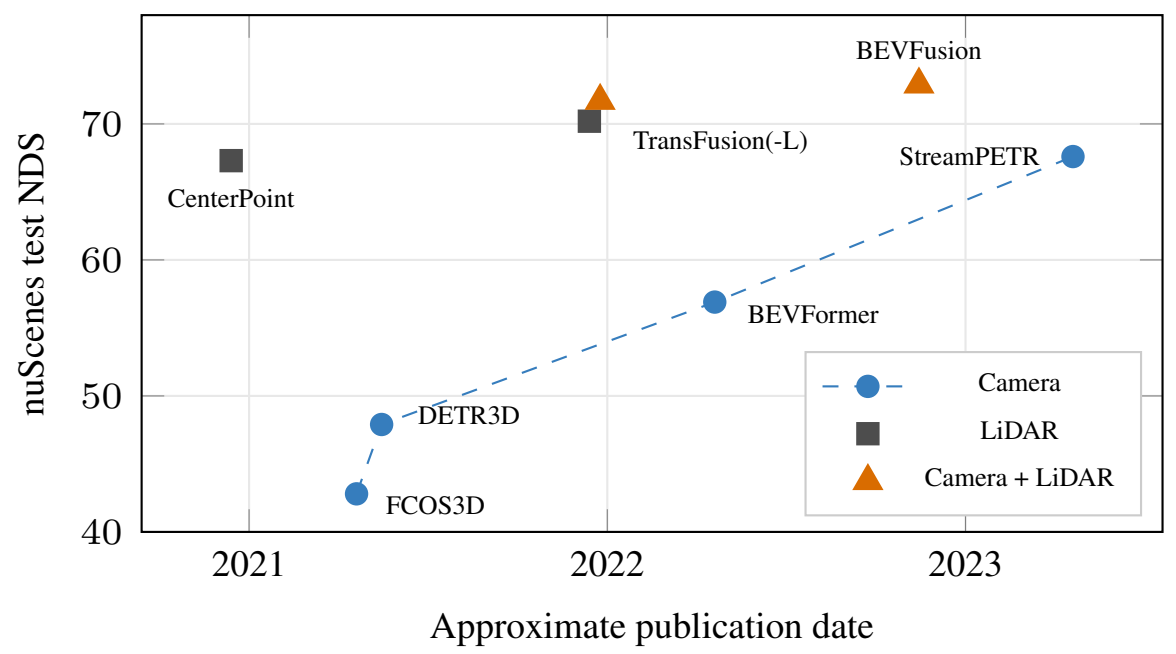

\noindent\textbf{Trend analysis.} On nuScenes test NDS (\cref{fig:nds}), camera-only detection rose from 42.8 (FCOS3D~\cite{wang2021fcos3d}, as reported in~\cite{wang2023streampetr}) and 47.9 (DETR3D~\cite{wang2021detr3d}) to 56.9 (BEVFormer~\cite{li2022bevformer}) and 67.6 (StreamPETR~\cite{wang2023streampetr}), slightly above LiDAR-only CenterPoint (67.3)~\cite{yin2021centerpoint}, while LiDAR (TransFusion-L, 70.2) and fusion (TransFusion 71.7~\cite{bai2022transfusion}; BEVFusion 72.9~\cite{liu2023bevfusion}) stay ahead. Backbones differ, so the trend is indicative; temporal state modeling drove much of the camera-only gain~\cite{li2022bevformer,wang2023streampetr}, supporting persistence (D3), while the residual gap argues for a metric geometric core.

\noindent\textbf{Results snapshot.} \Cref{tab:results} covers five further benchmarks. Occupancy heads improve on early BEV baselines by over 10 mIoU on Occ3D~\cite{tian2023occ3d,hou2024fastocc}; cooperation adds about 14 AP@0.5 on DAIR-V2X~\cite{hu2022where2comm}; diffusion-based planning reaches 88.1 PDMS on NAVSIM~\cite{liao2025diffusiondrive}; and on Bench2Drive, UniAD and VAD score below 50 while the best distilled baselines reach about 64~\cite{jia2024bench2drive}, so closed-loop robustness remains the bottleneck.

\begin{table}[t]
\centering
\caption{\textbf{Results snapshot} across \usr readouts, as reported in the original works (or in the cited work for re-evaluated baselines, marked $^\dagger$). C: camera; L: LiDAR. Higher is better except minADE.}
\label{tab:results}
\scriptsize
\setlength{\tabcolsep}{1.6pt}
\begin{tabular}{@{}l l c l r@{}}
\toprule
Benchmark & Method & Input & Metric & Score\\
\midrule
\multirow{4}{*}{\shortstack[l]{Occ3D-nuScenes\\(val)}} & BEVFormer$^\dagger$~\cite{tian2023occ3d} & C & mIoU & 26.88\\
 & CTF-Occ~\cite{tian2023occ3d} & C & mIoU & 28.53\\
 & SurroundOcc$^\dagger$~\cite{hou2024fastocc} & C & mIoU & 37.18\\
 & FastOcc~\cite{hou2024fastocc} & C & mIoU & 39.21\\
\midrule
WOMD (test) & MTR~\cite{shi2022mtr} & Tracks & mAP / minADE & 0.413 / 0.605\\
\midrule
\multirow{4}{*}{\shortstack[l]{DAIR-V2X\\(V2I)}} & No collaboration$^\dagger$~\cite{hu2022where2comm} & L & AP@0.5 / 0.7 & 50.0 / 43.6\\
 & Late fusion$^\dagger$~\cite{hu2022where2comm} & L & AP@0.5 / 0.7 & 53.1 / 37.9\\
 & V2X-ViT$^\dagger$~\cite{xu2022v2xvit} & L & AP@0.5 / 0.7 & 54.3 / 43.4\\
 & Where2comm~\cite{hu2022where2comm} & L & AP@0.5 / 0.7 & 63.7 / 48.9\\
\midrule
\multirow{3}{*}{NAVSIM navtest} & UniAD~\cite{hu2023uniad}  & C   & PDMS & 83.4\\
 & TransFuser~\cite{chitta2023transfuser}     & C+L & PDMS & 84.0\\
 & DiffusionDrive~\cite{liao2025diffusiondrive} & C+L & PDMS & 88.1\\
\midrule
\multirow{2}{*}{Bench2Drive} & VAD~\cite{jiang2023vad}  & C   & DS & 42.35\\
 & UniAD-Base~\cite{hu2023uniad}         & C   & DS & 45.81\\
\bottomrule
\end{tabular}
\end{table}

\noindent\textbf{Key findings.} We draw five findings.
\textbf{(F1)}~BEV and occupancy are the most shareable states: the cooperative methods surveyed fuse mainly in them~\cite{xu2022v2xvit,hu2022where2comm,yu2025univ2x}.
\textbf{(F2)}~Temporal state can matter as much as the sensor suite (\cref{fig:nds}).
\textbf{(F3)}~Cooperation pays off (\cref{tab:results}) but is rarely reported against bandwidth and latency together.
\textbf{(F4)}~Evaluation is moving to closed loop, where Bench2Drive driving scores stay well below 100~\cite{jia2024bench2drive}, as open-loop metrics can mislead~\cite{li2024egostatus}.
\textbf{(F5)}~The largest gap is data, not models: no dataset in \cref{tab:datasets} jointly covers infrastructure with aerial views or network-wide traces (\cref{fig:coverage}).

\section{USR-5: A Proposed Benchmark Protocol}
\label{sec:usr5}

Leaderboards cannot tell whether a state is \emph{shared}, so we propose \emph{USR-5}: five tests comparing a shared state with a reference on the same data (\cref{tab:usr5}). T1--T4 target D1--D4, T5 targets transfer (P2), and D5 is reported via bytes in T1.
\textbf{(T1)~Write}: accuracy after adding a platform or rig, plus bytes sent.
\textbf{(T2)~Read}: mean score of $\geq$3 heads $g_k$ on one frozen $\mathcal{S}_t$, each relative to a task-specific model (so $s_2^{\text{ref}}=1$).
\textbf{(T3)~Persistence}: accuracy on a revisit using prior state, also after map change~\cite{lambert2021tbv}.
\textbf{(T4)~Calibration}: $1-\text{ECE}$, with unknown-object recall~\cite{singh2026unknown} as a diagnostic.
\textbf{(T5)~Transfer}: accuracy on held-out cities, rigs, and conditions~\cite{wang2020traingermany}.

\noindent\textbf{Aggregate score.} Let $s_i>0$ be the shared state's score on test $i$ and $s_i^{\text{ref}}>0$ the reference's score under the same condition (higher is better). Then
\begin{equation}
\text{USR-5} = \Big(\textstyle\prod_{i=1}^{5}\min\!\big(s_i / s_i^{\text{ref}},\,2\big)\Big)^{1/5}.
\label{eq:usr5}
\end{equation}
References (\cref{tab:usr5}) should be the strongest published single-platform model or one matched in backbone and training budget, since a weak reference inflates the ratio. Values above 1 mean sharing helps; the geometric mean rewards balance and the cap bounds each test. As a worked example of T1 from published numbers, cooperative detection on DAIR-V2X reaches 63.7 AP@0.5 versus 50.0 without collaboration~\cite{hu2022where2comm}, so $s_1/s_1^{\text{ref}}=1.27$. USR-5 is an outlook, not a validated benchmark: no system has been scored on all five tests and only T1 can be filled from published numbers; its value now is to fix what to measure and against what, and scoring one full system is the first item of our agenda. Bytes and unknown-object recall stay diagnostics, as a reference may score zero on them.

\begin{table}[t]
\centering
\caption{\textbf{USR-5 tests}, the reference each score is divided by, and data that support them today.}
\label{tab:usr5}
\scriptsize
\setlength{\tabcolsep}{2.4pt}
\begin{tabularx}{\columnwidth}{@{}l l X@{}}
\toprule
Test & Reference $s_i^{\text{ref}}$ (same data) & Usable data today\\
\midrule
T1 Write       & Same model, ego platform only & DAIR-V2X, V2V4Real~\cite{yu2022dairv2x,xu2023v2v4real}\\
T2 Read        & Task-specific model per head & nuScenes, AV2~\cite{caesar2020nuscenes,wilson2021argoverse2}\\
T3 Persistence & Same model, empty prior state & TbV (static)~\cite{lambert2021tbv}\\
T4 Calibration & Single-platform model & Any labeled set\\
T5 Transfer    & Single-platform, same training & nuScenes, Waymo, ACDC~\cite{sun2020waymo,sakaridis2021acdc}\\
\bottomrule
\end{tabularx}
\end{table}

\section{Challenges and Research Agenda}
\label{sec:challenges}
\label{sec:future}

\Cref{tab:agenda} pairs each challenge with an open problem (P1--P6), directions, and measures.

\noindent\textbf{Transfer (P2).} Detectors lose substantial accuracy across countries, partly from differences in object size, sensors, and layout~\cite{wang2020traingermany}, and adverse conditions widen the gap~\cite{sakaridis2021acdc,paek2022kradar}. Foundation priors~\cite{radford2021clip,kirillov2023sam} transfer semantics, and 3D geometric foundation models~\cite{wang2025vggt,yang2024depthanything} begin to transfer geometry, but behavior remains city-specific and write-anywhere encoders for driving are largely unexplored.

\noindent\textbf{Uncertainty and open world (P4).} Calibrated aleatoric and epistemic uncertainty~\cite{kendall2017uncertainties} must propagate through fusion, latency, and pose error~\cite{xu2022v2xvit} to planning risk, and confidence can decide what to share~\cite{hu2022where2comm}. Open-set detection~\cite{singh2026unknown}, language priors~\cite{tian2024drivevlm}, and generative data~\cite{gao2024magicdrive} help with rare objects, but few urban benchmarks report unknown-object recall.

\noindent\textbf{Efficiency and persistence (P3).} Dense occupancy and video world models are costly and V2X bandwidth is limited, so a persistent city-scale state needs vectorized~\cite{liao2023maptr} or sparse states with change-aware updates.

\noindent\textbf{Evaluation validity (P5) and governance.} Neural simulators~\cite{yang2023unisim,yan2024streetgaussians} and world models~\cite{hu2023gaia1,zheng2024occworld} increasingly \emph{evaluate} planners without validated fidelity on rare events. A city-scale \usr also aggregates imagery and traces of people, so privacy and data ownership are prerequisites.

\noindent\textbf{Limitations of this survey.} The scorecards in \cref{tab:representations,tab:paradigms} are rubric-based judgments, not measurements; results in \cref{tab:results} come from different settings; and our coverage of urban computing is introductory.

\section{Conclusion}
\label{sec:conclusion}

This survey introduced \usr, a layered, georeferenced state with five desiderata, and related representations, architectures, datasets, and results to it. The field has shareable states and maturing closed-loop evaluation, but lacks joint multi-platform data and a measure of sharing, which USR-5 and the agenda target.

{
    \small
    \bibliographystyle{ieeenat_fullname}
    \bibliography{main}
}

\clearpage
\appendix
\providecommand{\mainref}[1]{#1 of the main paper}
\setcounter{section}{0}\setcounter{table}{0}\setcounter{figure}{0}\setcounter{equation}{0}
\renewcommand{\thesection}{S\arabic{section}}
\renewcommand{\thetable}{S\arabic{table}}
\renewcommand{\thefigure}{S\arabic{figure}}
\renewcommand{\theequation}{S\arabic{equation}}
\twocolumn[{\centering\Large\bf Supplementary Material\par\vspace{1.5em}}]

\section{Overview}
\label{sec:s-overview}

This supplement documents how the survey was assembled and makes its judgments checkable. It adds no new experimental results: every number below either appears in the main paper or is recomputed from it, and the USR-5 profiles in \cref{sec:s-usr5} are hypothetical illustrations of the formula, labeled as such. \Cref{tab:s-map} maps each item to the main-paper element it supports. The project repository contains the same tables as CSV files and a reference implementation of the USR-5 score (\cref{sec:s-files}).

\begin{table}[h]
\centering
\caption{\textbf{What each supplementary item supports.}}
\label{tab:s-map}
\scriptsize
\setlength{\tabcolsep}{3pt}
\begin{tabularx}{\columnwidth}{@{}l X l@{}}
\toprule
Item & Content & Supports (main paper)\\
\midrule
\cref{sec:s-protocol} & Selection criteria, verification, bibliometrics & Sec.~1 (methodology)\\
\cref{sec:s-rubric} & One-line justification per scorecard mark & Tab.~3\\
\cref{sec:s-data} & Platform flags; coverage recomputed & Tab.~6, Fig.~2\\
\cref{sec:s-prov} & Source of every reported number & Tab.~8, Fig.~3\\
\cref{sec:s-usr5} & Full USR-5 specification & Sec.~9, Eq.~3, Tab.~9\\
\cref{sec:s-agenda} & First experiments per open problem & Tab.~5, Sec.~10\\
\cref{sec:s-timeline} & Historical milestones & Sec.~1\\
\bottomrule
\end{tabularx}
\end{table}

\section{Literature Selection and Verification}
\label{sec:s-protocol}

\noindent\textbf{Type of review.} As stated in Sec.~1 of the main paper, this is a structured narrative survey, not a systematic review; we therefore report criteria and verification steps rather than screening counts.

\noindent\textbf{Seeds and expansion.} We started from the eleven related surveys compared in \mainref{Tab.~1}~\cite{li2024bevsurvey,xu2025occsurvey,han2023collabsurvey,huang2025v2xcp,chen2024e2esurvey,cui2024mllmsurvey,chen2024gssurvey,feng2025wmsurvey,tu2026dwmsurvey,zhang2024ufmsurvey,hu2026maead} and followed their citations backward and forward, together with the proceedings of the venues listed in Sec.~1 of the main paper, focusing on 2017--2026 plus earlier foundational work.

\noindent\textbf{Inclusion.} A work is included if it (a)~proposes or evaluates a representation used for urban perception, prediction, planning, or simulation; (b)~releases an urban dataset, simulator, or benchmark protocol; or (c)~is a survey that a reader of this one would need for depth on a sub-topic.

\noindent\textbf{Exclusion and selection.} We exclude work outside road and urban mobility, and cite the archival version of a work when one exists. Within a line of work we cite the first or most widely adopted method and the current representative, rather than every variant, so citations are representative rather than exhaustive (main paper, Sec.~10, Limitations).

\noindent\textbf{Verification.} Every one of the 94 bibliography entries was checked against the official proceedings, the publisher's page, or the arXiv record for title, author list, venue, and year. A preprint is cited as arXiv unless its venue could be confirmed, and a non-archival workshop paper is marked as such. Dataset scale figures in \mainref{Tab.~6} were checked against the original dataset papers, and every number in \mainref{Tab.~8} and \mainref{Fig.~3} was checked against the paper listed in \cref{tab:s-prov}.

\noindent\textbf{Bibliometrics.} \Cref{fig:s-years} and \cref{tab:s-cat} summarize the cited works. 74 of 94 (79\%) date from 2021--2026, and the venues span vision, machine learning, robotics and transportation, and data mining, reflecting the cross-community scope of the survey.

\def\yearcoords{(1,6) (2,2) (3,4) (4,2) (5,6) (6,10) (7,13) (8,16) (9,21) (10,10) (11,4)}
\def\yearlabels{{$\leq$2016},{2017},{2018},{2019},{2020},{2021},{2022},{2023},{2024},{2025},{2026}}
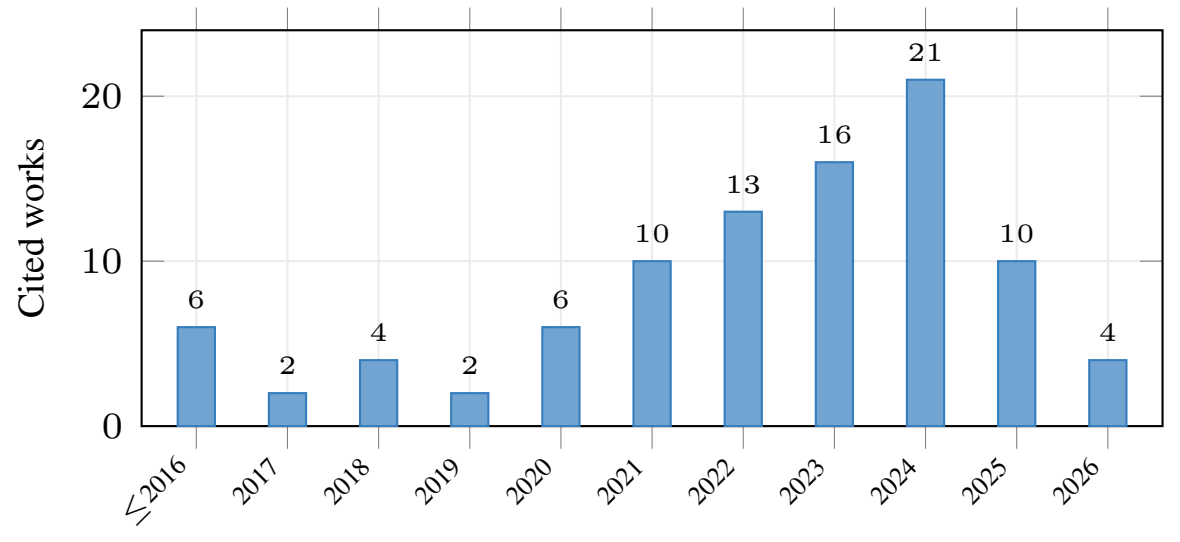
\begin{figure}[t]
\centering
\begin{tikzpicture}
\begin{axis}[
  width=\columnwidth, height=4.2cm, ybar, bar width=7pt,
  ymin=0, ymax=24, xmin=0.4, xmax=11.6,
  xtick={1,...,11}, xticklabels/.expanded={\yearlabels},
  x tick label style={font=\tiny, rotate=45, anchor=east},
  y tick label style={font=\scriptsize}, ylabel={Cited works}, label style={font=\scriptsize},
  nodes near coords, every node near coord/.append style={font=\tiny},
  grid=major, grid style={gray!15}]
\addplot[fill=wacvblue!70, draw=wacvblue] coordinates {\yearcoords};
\end{axis}
\end{tikzpicture}
\caption{\textbf{Cited works by publication year} (all 94 bibliography entries).}
\label{fig:s-years}
\end{figure}

\begin{table}[t]
\centering
\caption{\textbf{Cited works by venue type.}}
\label{tab:s-cat}
\scriptsize
\begin{tabular}{@{}l r@{}}
\toprule
Venue type & Works\\
\midrule
Computer vision conference / workshop & 45\\
Machine learning / AI venue & 22\\
Journal (IEEE, ACM, Elsevier, Springer) & 10\\
Preprint (arXiv) & 7\\
Robotics / ITS conference & 6\\
Data mining / GIS / urban computing & 4\\
\midrule
Total & 94\\
\bottomrule
\end{tabular}
\end{table}

\section{Justification of the Representation Scorecard}
\label{sec:s-rubric}

\Cref{tab:s-rubric} gives a one-line reason for each of the 40 marks in \mainref{Tab.~3}, applying the rubric in its caption: D1 \cmark\ if any posed platform can write without retraining the state; D2 \cmark\ if three or more task heads have been shown on one state; D3 \cmark\ if the state is map-anchored and reused across traversals; D4 \cmark\ if per-element uncertainty is output; D5 \cmark\ if memory or bandwidth grows sub-linearly with area or the state is sparse or vectorized. \pmark\ means met by some published variants, \xmark\ not met in typical practice. These are our reading of typical practice, not measurements; the reasons make each judgment explicit so that it can be disputed.

\begin{table*}[t]
\centering
\caption{\textbf{Reason for each mark in Tab.~3 of the main paper.} \cmark: met; \pmark: met by some variants; \xmark: not met in typical practice.}
\label{tab:s-rubric}
\scriptsize
\setlength{\tabcolsep}{3pt}
\renewcommand{\arraystretch}{1.05}
\begin{tabularx}{\textwidth}{@{}l c c X@{}}
\toprule
Family & Des. & Mark & Reason\\
\midrule
\multirow{5}{*}{\parbox{1.7cm}{BEV grid}} & D1 & \cmark & Any posed camera, LiDAR, or radar can be projected or splatted onto the ground plane~\cite{philion2020lss,liu2023bevfusion}; V2X methods fuse BEV maps from other agents~\cite{xu2022v2xvit}.\\
 & D2 & \cmark & Detection, map segmentation, and motion heads share one BEV feature map~\cite{li2022bevformer,hu2023uniad}.\\
 & D3 & \pmark & The grid is ego-centric with a fixed extent; temporal attention keeps only a short history~\cite{li2022bevformer}.\\
 & D4 & \pmark & Most heads output scores, not calibrated per-cell uncertainty; confidence maps exist in some variants~\cite{hu2022where2comm}.\\
 & D5 & \pmark & Dense grids grow with area; confidence-driven sparse messages reduce bandwidth~\cite{hu2022where2comm}.\\
\midrule
\multirow{5}{*}{\parbox{1.7cm}{3D occupancy}} & D1 & \cmark & Voxels in a shared frame can be filled from any posed sensor~\cite{tian2023occ3d,wei2023surroundocc}.\\
 & D2 & \cmark & Semantic occupancy supports segmentation, planning, and forecasting~\cite{zheng2024occworld}.\\
 & D3 & \pmark & Predictions are usually per frame or short-history, not accumulated across traversals.\\
 & D4 & \pmark & Visibility masks~\cite{tian2023occ3d} and probabilistic variants exist, but per-voxel calibrated uncertainty is uncommon.\\
 & D5 & \xmark & Memory is cubic in extent; Gaussian and sparse variants~\cite{huang2024gaussianformer} mitigate but are not yet typical.\\
\midrule
\multirow{5}{*}{\parbox{1.7cm}{Vectorized map}} & D1 & \cmark & Polylines live in a map frame and can be contributed by any platform that observes lanes~\cite{liao2023maptr}.\\
 & D2 & \pmark & Vectors serve planning and prediction~\cite{jiang2023vad} but carry no appearance or dynamic agents.\\
 & D3 & \cmark & Map elements persist across trips and form the static layer.\\
 & D4 & \pmark & Few online mappers output per-element uncertainty.\\
 & D5 & \cmark & A few points per element; compact for storage and transmission.\\
\midrule
\multirow{5}{*}{\parbox{1.7cm}{Neural field / 3DGS}} & D1 & \pmark & Writing requires posed images and usually per-scene optimization~\cite{mildenhall2020nerf,kerbl20233dgs}; feed-forward variants are emerging~\cite{yang2025storm}.\\
 & D2 & \pmark & Mainly rendering and simulation~\cite{yang2023unisim}; few task heads read the field directly.\\
 & D3 & \cmark & A reconstructed scene is a persistent, georeferenceable asset~\cite{tancik2022blocknerf}.\\
 & D4 & \xmark & Typical pipelines output no uncertainty.\\
 & D5 & \pmark & Block decomposition scales to city blocks~\cite{tancik2022blocknerf}, but memory and fitting cost remain high.\\
\midrule
\multirow{5}{*}{\parbox{1.7cm}{Scene graph}} & D1 & \cmark & Nodes for agents, lanes, and signals can come from any platform's detections.\\
 & D2 & \cmark & Graph-structured QA drives perception, prediction, and planning answers~\cite{sima2024drivelm}.\\
 & D3 & \pmark & Graphs are usually built per frame or per clip.\\
 & D4 & \xmark & Edges and nodes rarely carry calibrated uncertainty.\\
 & D5 & \cmark & Symbolic and compact.\\
\midrule
\multirow{5}{*}{\parbox{1.7cm}{Language tokens}} & D1 & \pmark & Writing needs a vision-language model per platform; non-image modalities are rarely supported~\cite{tian2024drivevlm}.\\
 & D2 & \cmark & Description, QA, and planning share the same tokens~\cite{hwang2024emma,sima2024drivelm}.\\
 & D3 & \xmark & Text is neither georeferenced nor persistent across trips.\\
 & D4 & \xmark & Generated text carries no calibrated uncertainty and can hallucinate~\cite{cui2024mllmsurvey}.\\
 & D5 & \cmark & Tokens are compact to store and send (latency is the cost, see main Tab. 4).\\
\midrule
\multirow{5}{*}{\parbox{1.7cm}{World-model latent}} & D1 & \xmark & The latent is specific to one trained model and ego viewpoint~\cite{hu2023gaia1}.\\
 & D2 & \pmark & Generation and planning read it; other perception heads rarely do~\cite{zheng2024occworld}.\\
 & D3 & \xmark & Latents are not map-anchored or reused across traversals.\\
 & D4 & \pmark & Sampling yields a distribution over futures, but not calibrated per-element uncertainty.\\
 & D5 & \pmark & The latent is compact, but video generation is expensive~\cite{hu2023gaia1}.\\
\midrule
\multirow{5}{*}{\parbox{1.7cm}{Road-network graph}} & D1 & \pmark & Loop detectors and map-matched traces write it~\cite{li2018dcrnn,newson2009mapmatching}; onboard perception rarely does.\\
 & D2 & \pmark & Speed, flow, and demand forecasts share it~\cite{yu2018stgcn,wu2019graphwavenet}, but not perception tasks.\\
 & D3 & \cmark & The graph is fixed to the road network and persists.\\
 & D4 & \pmark & Probabilistic forecasting is possible but not the norm in the cited models.\\
 & D5 & \cmark & Signals on nodes and edges scale with network size, not area.\\
\midrule
\end{tabularx}
\end{table*}

\section{Dataset Platform Flags and Coverage}
\label{sec:s-data}

\Cref{tab:s-flags} records, for each of the 26 datasets in \mainref{Tab.~6}, which platforms it observes jointly, whether it is synthetic, and which USR-5 tests it could support. Ego means street-level vehicle sensors (MatrixCity counts because it renders street-level views); Infrastructure means roadside sensors; Aerial means drone or satellite imagery; Traces means GNSS trajectories or loop detectors (BDD100K counts through its dashcam GPS). \Cref{tab:s-cov} recomputes the off-diagonal cells of \mainref{Fig.~2} from these flags; all ten cells match the figure, and \texttt{code/check\_coverage.py} in the repository reproduces them from the CSV. Candidate T1 sets contain at least two writing platforms, T2 sets have three or more annotated tasks, and T5 sets span several cities or conditions. No dataset in the table supports T3 (persistence across revisits with change labels); the map-change dataset TbV~\cite{lambert2021tbv} is the closest current option.

\begin{table}[t]
\centering
\caption{\textbf{Platform flags} for the datasets in Tab.~6 of the main paper. E: ego; I: infrastructure; A: aerial; Tr: traces; MV: multiple vehicles. Last three columns: candidate data for USR-5 tests T1, T2, T5.}
\label{tab:s-flags}
\scriptsize
\setlength{\tabcolsep}{2.4pt}
\begin{tabular}{@{}l c c c c c c c c c@{}}
\toprule
Dataset & E & I & A & Tr & MV & Data & T1 & T2 & T5\\
\midrule
KITTI~\cite{geiger2012kitti} & \cmark &  &  &  &  & real &  &  & \\
Cityscapes~\cite{cordts2016cityscapes} & \cmark &  &  &  &  & real &  &  & \cmark\\
nuScenes~\cite{caesar2020nuscenes} & \cmark &  &  &  &  & real &  & \cmark & \cmark\\
Waymo Open~\cite{sun2020waymo} & \cmark &  &  &  &  & real &  & \cmark & \cmark\\
BDD100K~\cite{yu2020bdd100k} & \cmark &  &  & \cmark &  & real &  & \cmark & \cmark\\
WOMD~\cite{ettinger2021womd} & \cmark &  &  &  &  & real &  &  & \cmark\\
Argoverse 2~\cite{wilson2021argoverse2} & \cmark &  &  &  &  & real &  & \cmark & \cmark\\
Occ3D~\cite{tian2023occ3d} & \cmark &  &  &  &  & real &  &  & \\
ACDC~\cite{sakaridis2021acdc} & \cmark &  &  &  &  & real &  &  & \cmark\\
K-Radar~\cite{paek2022kradar} & \cmark &  &  &  &  & real &  &  & \cmark\\
OPV2V~\cite{xu2022opv2v} & \cmark &  &  &  & \cmark & sim & \cmark &  & \\
DAIR-V2X~\cite{yu2022dairv2x} & \cmark & \cmark &  &  &  & real & \cmark &  & \\
V2V4Real~\cite{xu2023v2v4real} & \cmark &  &  &  & \cmark & real & \cmark &  & \\
V2X-Seq~\cite{yu2023v2xseq} & \cmark & \cmark &  &  &  & real & \cmark &  & \\
TUMTraf V2X~\cite{zimmer2024tumtraf} & \cmark & \cmark &  &  &  & real & \cmark &  & \\
highD~\cite{krajewski2018highd} &  &  & \cmark &  &  & real &  &  & \\
VisDrone~\cite{zhu2022visdrone} &  &  & \cmark &  &  & real &  &  & \\
DeepGlobe~\cite{demir2018deepglobe} &  &  & \cmark &  &  & real &  &  & \\
MatrixCity~\cite{li2023matrixcity} & \cmark &  & \cmark &  &  & sim &  &  & \\
GeoLife~\cite{zheng2010geolife} &  &  &  & \cmark &  & real &  &  & \\
METR-LA~\cite{li2018dcrnn} &  &  &  & \cmark &  & real &  &  & \\
CARLA~\cite{dosovitskiy2017carla} & \cmark &  &  &  &  & sim &  &  & \\
nuPlan~\cite{caesar2021nuplan} & \cmark &  &  &  &  & real &  &  & \cmark\\
NAVSIM~\cite{dauner2024navsim} & \cmark &  &  &  &  & real &  &  & \\
Bench2Drive~\cite{jia2024bench2drive} & \cmark &  &  &  &  & sim &  &  & \\
DriveLM~\cite{sima2024drivelm} & \cmark &  &  &  &  & real &  &  & \\
\bottomrule
\end{tabular}
\end{table}

\begin{table}[t]
\centering
\caption{\textbf{Off-diagonal coverage of Fig.~2 recomputed} from \cref{tab:s-flags}. Multi-vehicle pairing (OPV2V, V2V4Real) is within the ego platform.}
\label{tab:s-cov}
\scriptsize
\setlength{\tabcolsep}{3pt}
\begin{tabularx}{\columnwidth}{@{}l c X@{}}
\toprule
Pair & Count & Datasets\\
\midrule
Ego--Infra & 3 & DAIR-V2X, TUMTraf V2X, V2X-Seq\\
Ego--Aerial & 1 & MatrixCity\\
Ego--Traces & 1 & BDD100K\\
Infra--Aerial & 0 & \textcolor{red!70!black}{none}\\
Infra--Traces & 0 & \textcolor{red!70!black}{none}\\
Aerial--Traces & 0 & \textcolor{red!70!black}{none}\\
\bottomrule
\end{tabularx}
\end{table}

\section{Provenance of Reported Numbers}
\label{sec:s-prov}

\Cref{tab:s-prov} lists every number in \mainref{Tab.~8} and \mainref{Fig.~3} with the paper it was taken from. \emph{Original} means the method's own paper; \emph{re-evaluated} means a later paper re-ran the baseline under its protocol; \emph{as listed} means the value is quoted from a comparison table in a later paper; \emph{benchmark baseline} means it was reported by the benchmark's authors. Values were copied without conversion or rounding beyond the source.

\noindent\textbf{Comparability.} Rows within one benchmark share a split and metric, but methods differ in backbone, input resolution, training data, and test-time settings, so differences are indicative only (main paper, Sec.~10). The nuScenes series in \mainref{Fig.~3} mixes backbones; publication dates on its x-axis are approximate. For DAIR-V2X, all four rows come from one paper under one protocol~\cite{hu2022where2comm}, which is why that benchmark was chosen for the USR-5 worked example.

\begin{table*}[t]
\centering
\caption{\textbf{Source of each number} in Tab.~8 and Fig.~3 of the main paper. C: camera; L: LiDAR.}
\label{tab:s-prov}
\scriptsize
\setlength{\tabcolsep}{3.2pt}
\begin{tabular}{@{}l l l l l l l l l@{}}
\toprule
Benchmark & Split & Method & Input & Metric & Value & Reported in & Status & Used in\\
\midrule
Occ3D-nuScenes & val & BEVFormer & C & mIoU & 26.88 & \cite{tian2023occ3d} & re-evaluated & Tab. 8\\
Occ3D-nuScenes & val & CTF-Occ & C & mIoU & 28.53 & \cite{tian2023occ3d} & original & Tab. 8\\
Occ3D-nuScenes & val & SurroundOcc & C & mIoU & 37.18 & \cite{hou2024fastocc} & re-evaluated & Tab. 8\\
Occ3D-nuScenes & val & FastOcc & C & mIoU & 39.21 & \cite{hou2024fastocc} & original & Tab. 8\\
WOMD & test & MTR & Tracks & mAP / minADE & 0.413 / 0.605 & \cite{shi2022mtr} & original & Tab. 8\\
DAIR-V2X (V2I) & -- & No collaboration & L & AP@0.5 / 0.7 & 50.0 / 43.6 & \cite{hu2022where2comm} & re-evaluated & Tab. 8, USR-5 T1\\
DAIR-V2X (V2I) & -- & Late fusion & L & AP@0.5 / 0.7 & 53.1 / 37.9 & \cite{hu2022where2comm} & re-evaluated & Tab. 8\\
DAIR-V2X (V2I) & -- & V2X-ViT & L & AP@0.5 / 0.7 & 54.3 / 43.4 & \cite{hu2022where2comm} & re-evaluated & Tab. 8\\
DAIR-V2X (V2I) & -- & Where2comm & L & AP@0.5 / 0.7 & 63.7 / 48.9 & \cite{hu2022where2comm} & original & Tab. 8, USR-5 T1\\
NAVSIM & navtest & UniAD & C & PDMS & 83.4 & \cite{liao2025diffusiondrive} & as listed & Tab. 8\\
NAVSIM & navtest & TransFuser & C+L & PDMS & 84.0 & \cite{liao2025diffusiondrive} & as listed & Tab. 8\\
NAVSIM & navtest & DiffusionDrive & C+L & PDMS & 88.1 & \cite{liao2025diffusiondrive} & original & Tab. 8\\
Bench2Drive & closed loop & VAD & C & Driving score & 42.35 & \cite{jia2024bench2drive} & benchmark baseline & Tab. 8\\
Bench2Drive & closed loop & UniAD-Base & C & Driving score & 45.81 & \cite{jia2024bench2drive} & benchmark baseline & Tab. 8\\
nuScenes & test & FCOS3D & C & NDS & 42.8 & \cite{wang2023streampetr} & as listed & Fig. 3\\
nuScenes & test & DETR3D & C & NDS & 47.9 & \cite{wang2021detr3d} & original & Fig. 3\\
nuScenes & test & BEVFormer & C & NDS & 56.9 & \cite{li2022bevformer} & original & Fig. 3\\
nuScenes & test & StreamPETR & C & NDS & 67.6 & \cite{wang2023streampetr} & original & Fig. 3\\
nuScenes & test & CenterPoint & L & NDS & 67.3 & \cite{yin2021centerpoint} & original & Fig. 3\\
nuScenes & test & TransFusion-L & L & NDS & 70.2 & \cite{bai2022transfusion} & original & Fig. 3\\
nuScenes & test & TransFusion & C+L & NDS & 71.7 & \cite{bai2022transfusion} & original & Fig. 3\\
nuScenes & test & BEVFusion & C+L & NDS & 72.9 & \cite{liu2023bevfusion} & original & Fig. 3\\
\bottomrule
\end{tabular}
\end{table*}

\section{USR-5: Full Specification}
\label{sec:s-usr5}

\noindent\textbf{Per-test protocol.} \Cref{tab:s-usr5} expands \mainref{Tab.~9}. In every test the shared state and the reference use the same data split, and the reference should be the strongest published single-platform, single-task model, or a baseline matched in backbone and training budget; a weak reference inflates the ratio.

\begin{table*}[t]
\centering
\caption{\textbf{USR-5 test specification.} $s_i$: shared-state score; $s_i^{\text{ref}}$: reference score on the same split. Diagnostics are reported alongside but do not enter Eq.~3 of the main paper.}
\label{tab:s-usr5}
\scriptsize
\setlength{\tabcolsep}{3pt}
\begin{tabularx}{\textwidth}{@{}l >{\raggedright\arraybackslash}X >{\raggedright\arraybackslash}X l >{\raggedright\arraybackslash}X >{\raggedright\arraybackslash}X@{}}
\toprule
Test & Setup & Reference & Score $s_i$ & Diagnostics & Candidate data\\
\midrule
T1 Write & Add a platform or sensor rig (\eg, roadside LiDAR) that writes into the shared state & Same model, ego platform only & Task accuracy (\eg, AP) & Bytes per frame; latency & DAIR-V2X~\cite{yu2022dairv2x}, V2V4Real~\cite{xu2023v2v4real}, OPV2V (sim)~\cite{xu2022opv2v}\\
T2 Read & Freeze $\mathcal{S}_t$; train $\geq$3 light heads (\eg, detection, map segmentation, motion) & A specialist per task on the same input & Mean of head/specialist ratios ($s_2^{\text{ref}}=1$) & Per-head ratios & nuScenes~\cite{caesar2020nuscenes}, AV2~\cite{wilson2021argoverse2}\\
T3 Persistence & Revisit a location with prior state loaded, including a map-change subset & Same model with an empty prior state & Accuracy on revisits & Accuracy on changed regions; staleness & TbV~\cite{lambert2021tbv} (static layer only)\\
T4 Calibration & Per-element confidence from the shared state & Single-platform model & $1-\text{ECE}$ & Unknown-object recall; ECE binning & Any labeled set\\
T5 Transfer & Train on source cities, rigs, or conditions; test on held-out ones & Single-platform model trained identically & Held-out accuracy & Drop from in-domain accuracy & nuScenes, Waymo~\cite{sun2020waymo}, ACDC~\cite{sakaridis2021acdc}\\
\bottomrule
\end{tabularx}
\end{table*}

\noindent\textbf{Properties of the aggregate.} With $r_i=\min(s_i/s_i^{\text{ref}},2)$ and USR-5 $=(\prod_i r_i)^{1/5}$ (main paper, Eq.~3):
(i)~each factor is a ratio, so tests with different metrics combine without unit conversion;
(ii)~the score lies in $(0,2]$, and 1 means parity with the reference;
(iii)~the geometric mean prevents a failed test from being offset by gains elsewhere;
(iv)~the cap prevents one extreme gain from dominating;
(v)~the score is non-decreasing in every $s_i$;
(vi)~if a test is missing, USR-5 is undefined; a partial score over $k$ tests should be reported as USR-$k$ with the tests listed.
Bytes (T1) and unknown-object recall (T4) stay diagnostics because a reference may score zero on them, which would make a ratio undefined.

\noindent\textbf{Illustration (hypothetical, not results).} \Cref{tab:s-agg} applies three aggregation rules to three invented ratio profiles. An arithmetic mean ranks profile C (four gains, one failed test) equal to the balanced profile A; the geometric mean ranks it below parity. Profile B (one large gain, small losses elsewhere) scores highest under the arithmetic mean, but the cap brings it close to A.

\begin{table}[t]
\centering
\caption{\textbf{Aggregation rules on hypothetical ratio profiles} $(r_1,\dots,r_5)$, computed by the included script. These are not measurements.}
\label{tab:s-agg}
\scriptsize
\setlength{\tabcolsep}{3pt}
\begin{tabular}{@{}l l c c c@{}}
\toprule
Profile & Ratios & Arith. & Geo. & Geo.+cap\\
\midrule
A balanced & 1.1, 1.1, 1.1, 1.1, 1.1 & 1.100 & 1.100 & 1.100\\
B one large gain & 2.5, 0.9, 0.9, 0.9, 0.9 & 1.220 & 1.104 & 1.056\\
C one failure & 1.3, 1.3, 1.3, 1.3, 0.3 & 1.100 & 0.970 & 0.970\\
\bottomrule
\end{tabular}
\end{table}

\noindent\textbf{Worked example and reporting template.} \Cref{tab:s-report} is the form we propose for reporting USR-5, filled in only where published numbers exist. On DAIR-V2X, Where2comm reaches 63.7 AP@0.5 and 48.9 AP@0.7 against 50.0 and 43.6 without collaboration~\cite{hu2022where2comm}, giving T1 ratios of 1.27 and 1.12; the stricter threshold yields the smaller gain, so the IoU threshold must be reported with the score. No published system covers the remaining tests, so the aggregate is not yet defined for any method.

\begin{table}[t]
\centering
\caption{\textbf{USR-5 reporting template}, filled with the only published case (T1). --: not yet reported for any system.}
\label{tab:s-report}
\scriptsize
\setlength{\tabcolsep}{2.6pt}
\begin{tabular}{@{}l l c c c l@{}}
\toprule
Test & Data / metric & $s_i$ & $s_i^{\text{ref}}$ & $r_i$ & Diagnostics\\
\midrule
T1 & DAIR-V2X, AP@0.5 & 63.7 & 50.0 & 1.27 & bytes: report\\
T1 & DAIR-V2X, AP@0.7 & 48.9 & 43.6 & 1.12 & bytes: report\\
T2 & -- & -- & 1 & -- & per-head ratios\\
T3 & -- & -- & -- & -- & changed regions\\
T4 & -- & -- & -- & -- & unknown recall\\
T5 & -- & -- & -- & -- & in-domain drop\\
\midrule
\multicolumn{4}{@{}l}{USR-5} & \multicolumn{2}{l}{undefined (1 of 5 tests)}\\
\bottomrule
\end{tabular}
\end{table}

\noindent\textbf{Known limitations.} The equal weighting and the cap of 2 are design choices, not derived quantities; ECE depends on the binning scheme, which must be reported; T3 data currently cover only static map change; and the score depends on the strength of the reference, which is why the reference model must be named. We present USR-5 as a proposal to be refined by the community.

\section{First Experiments for the Open Problems}
\label{sec:s-agenda}

For each open problem in \mainref{Tab.~5}, a first experiment that existing data already allow:

\noindent\textbf{P1 Joint multi-platform data.} Co-register the intersections of an existing V2I dataset~\cite{yu2022dairv2x,zimmer2024tumtraf} with aerial imagery of the same sites, using cross-view matching~\cite{zhu2021vigor,zhu2022transgeo}, and report T1 with the aerial view as an additional writer.

\noindent\textbf{P2 Write-anywhere encoders.} Train on one nuScenes city and test on the other, in the spirit of the cross-country study of~\cite{wang2020traingermany}, with and without a geometric foundation backbone~\cite{wang2025vggt,yang2024depthanything}; report T5.

\noindent\textbf{P3 Persistent city-scale state.} On TbV~\cite{lambert2021tbv}, compare a stored vectorized map~\cite{liao2023maptr} updated only where change is detected against re-estimation every frame; report accuracy, bytes, and staleness.

\noindent\textbf{P4 Calibrated uncertainty.} Measure ECE of a cooperative detector~\cite{xu2022v2xvit,hu2022where2comm} as pose noise and latency are injected, separating aleatoric and epistemic terms~\cite{kendall2017uncertainties}; report T4.

\noindent\textbf{P5 Trustworthy generative evaluation.} Rank the same set of planners in a neural simulator~\cite{yang2023unisim} and in pseudo-simulation on real logs~\cite{dauner2024navsim}, and report the rank correlation between the two.

\noindent\textbf{P6 Bridging AV and urban computing.} Attach road-graph traffic signals~\cite{li2018dcrnn} to vectorized lane maps~\cite{liao2023maptr} in one city and test whether congestion features improve motion forecasting.

\section{Historical Timeline}
\label{sec:s-timeline}

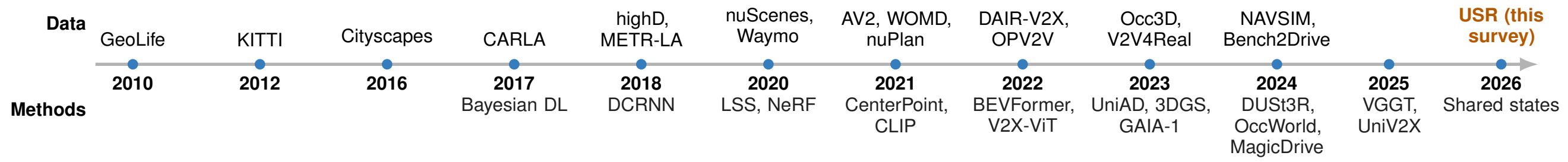
\begin{figure*}[t]
\centering
\resizebox{\textwidth}{!}{%
\begin{tikzpicture}[font=\sffamily\scriptsize]
\draw[very thick, gray!60, -{Latex[length=2.5mm]}] (-0.3,0) -- (19.0,0);
\foreach \y/\x in {2010/0.2,2012/1.9,2016/3.6,2017/5.3,2018/7.0,2020/8.7,2021/10.4,2022/12.1,2023/13.8,2024/15.5,2025/17.0,2026/18.5}{
  \fill[wacvblue] (\x,0) circle (2pt);
  \node[below=1pt, font=\sffamily\bfseries\scriptsize] at (\x,0) {\y};}
\node[left, font=\sffamily\bfseries\scriptsize] at (-0.3,0.55) {Data};
\node[left, font=\sffamily\bfseries\scriptsize] at (-0.3,-0.6) {Methods};
\foreach \x/\t in {0.2/GeoLife,1.9/KITTI,3.6/Cityscapes,5.3/CARLA,7.0/{highD, METR-LA},8.7/{nuScenes, Waymo},10.4/{AV2, WOMD, nuPlan},12.1/{DAIR-V2X, OPV2V},13.8/{Occ3D, V2V4Real},15.5/{NAVSIM, Bench2Drive}}
  \node[above, align=center, text width=1.6cm] at (\x,0.12) {\t};
\node[above, align=center, text width=1.6cm, orange!70!black, font=\sffamily\bfseries\scriptsize] at (18.5,0.12) {USR (this survey)};
\foreach \x/\t in {5.3/{Bayesian DL},7.0/DCRNN,8.7/{LSS, NeRF},10.4/{CenterPoint, CLIP},12.1/{BEVFormer, V2X-ViT},13.8/{UniAD, 3DGS, GAIA-1},15.5/{DUSt3R, OccWorld, MagicDrive},17.0/{VGGT, UniV2X},18.5/{Shared states}}
  \node[below, align=center, text width=1.6cm, gray!25!black] at (\x,-0.32) {\t};
\end{tikzpicture}}
\caption{\textbf{Milestones toward urban state representations.} Datasets and benchmarks (top) and methods (bottom), moving from single-sensor perception to cooperative, generative, and language-grounded representations.}
\label{fig:timeline}
\end{figure*}

\Cref{fig:timeline} places the main datasets and methods of the survey on a time axis, moving from single-sensor perception to cooperative, generative, and language-grounded representations.

\section{Data and Code}
\label{sec:s-files}

The project repository, \url{https://github.com/binayksingh/Awesome-Urban-State-Representation}, contains this material and:
\begin{itemize}\setlength\itemsep{0pt}
\item \texttt{data/datasets\_extended.csv}: \mainref{Tab.~6} with the flags of \cref{tab:s-flags};
\item \texttt{data/coverage\_matrix.csv}: all ten cells of \mainref{Fig.~2} with dataset names;
\item \texttt{data/results\_provenance.csv}: \cref{tab:s-prov};
\item \texttt{data/representation\_rubric.csv}: \cref{tab:s-rubric};
\item \texttt{code/check\_coverage.py}: recomputes \mainref{Fig.~2} from the dataset CSV;
\item \texttt{code/usr5.py}: reference implementation of Eq.~3, the worked example, and \cref{tab:s-agg} (Python 3, standard library only).
\end{itemize}

\end{document}